\documentclass[12pt,a4paper]{article}

\usepackage[pdftex]{graphicx}
\usepackage{natbib, color, setspace}
\usepackage[title]{appendix}
\usepackage{amsfonts, amssymb, amsmath}
\RequirePackage[colorlinks,citecolor=blue,urlcolor=blue]{hyperref}

\graphicspath{{figures/}}

\title{\fontsize{16}{12pt} \textbf{Functional Data Analysis and Visualisation of Three-dimensional Surface Shape}}

\author{Stanislav Katina \\
\normalsize Institute of Mathematics \& Statistics \\ \normalsize Mazaryk University, \normalsize Brno, Czech Republic
\and
Liberty Vittert \\
\normalsize Olin Business School \\ \normalsize Washington University in St.\ Louis, USA
\and
Adrian W.~Bowman \\
\normalsize School of Mathematics \& Statistics \\
\normalsize The University of Glasgow, U.K.\\
}

\begin{document}

\maketitle

\centerline{\textbf{Summary}}

{\small \noindent 
The advent of high resolution imaging has made data on surface shape widespread.   Methods for the analysis of shape based on landmarks are well established but high resolution data require a functional approach.  The starting point is a systematic and consistent description of each surface shape.  Three innovative forms of analysis are then introduced.  The first uses surface integration to address issues of registration, principal component analysis and the measurement of asymmetry, all in functional form.  Computational issues are handled through discrete approximations to integrals, based in this case on appropriate surface area weighted sums.  The second innovation is to focus on sub-spaces where interesting behaviour such as group differences are exhibited, rather than on individual principal components.  The third innovation concerns the comparison of individual shapes with a relevant control set, where the concept of a normal range is extended to the highly multivariate setting of surface shape.  This has particularly strong applications to medical contexts where the assessment of individual patients is very important.  All of these ideas are developed and illustrated in the important context of human facial shape, with a strong emphasis on the effective visual communication of effects of interest.
}

\


\newpage


\section{Introduction}
\label{sec:introduction}

\begin{figure}
\includegraphics[width = 0.3\textwidth]{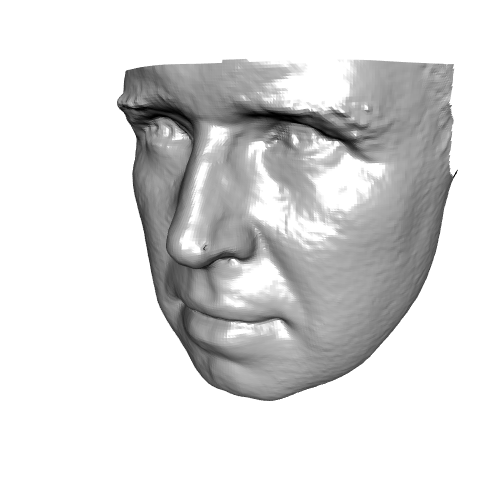}
\includegraphics[width = 0.3\textwidth]{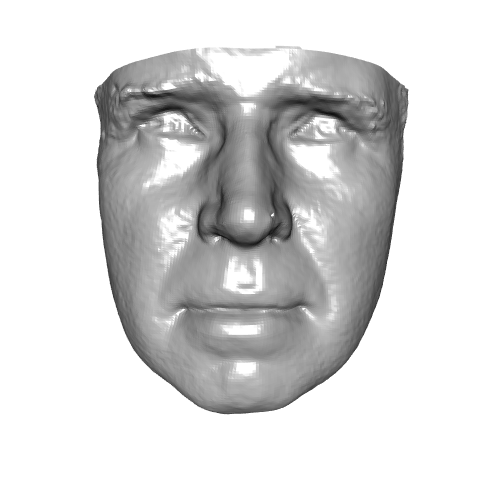}
\includegraphics[width = 0.3\textwidth]{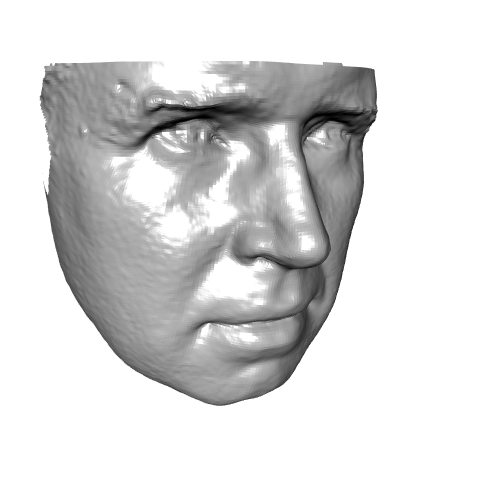}
\caption{An example of a 3D facial image, at different orientations.}
\label{fig:example}
\end{figure}

Statistical shape analysis is a research topic which has seen very substantial growth and development in recent years.  Early work in this area focused on representations of shape through carefully chosen landmarks, as point locations with an interpretation which corresponds across different shapes.  \citet{dryden-2016-book} provide a very comprehensive description of methods for the analysis of landmarks, but the later chapters of the book also indicate the much wider array of data types which are becoming available, driven by rapid advances in imaging technology.  A particular example is the increasing availability of sensors which employ techniques such as laser scanning or stereo-photogrammetry to create high-resolution data on surface shape in three dimensions.  This has a very wide variety of applications and it is the focus of the present paper.  Figure~\ref{fig:example} shows an image of a human face as an example of the kind of 3D surface data which is now easily obtainable.

Single instances of 3D surface data can be displayed in a variety of ways; in particular, the \texttt{rgl} package \citep{rgl-2019} is an indispensable tool for those from the R \citep{R} community, as it provides access to the \textit{OpenGL} industry-standard tools for 3D display.  However, the effective display of patterns and variation in collections of 3D objects is more challenging.  \cite{bowman-2006-jcgs} gave some discussion of this for 3D points and curves, but the aim of the present paper is to provide new tools for the modeling and visualisation of samples of 3D surface data.

The starting point is a description of an individual surface which has a consistent meaning across all the surfaces in the dataset.  This can be approached in different ways and the particular method adopted here is described in Section~\ref{sec:facial-model}.  Some obvious issues of analysis then commonly arise.  These include the need for methods to:
\begin{itemize}
\item   register the surfaces to a common co-ordinate system;
\item   characterise the variation present in a sample of surfaces;
\item   compare surface shapes across groups;
\item   assess the surface shape of an individual against a relevant control set.
\end{itemize}
These problems are tackled here from a functional perspective.  Adaptations of standard methods of Procrustes analysis are introduced in Section~\ref{sec:facial-model}, using a metric based on an approximated surface integral rather than individual point locations.  Non-linear registration through warping is also described as a means of displaying the results of analysis at higher resolution, for visual effect.  Basic methods of visualising surface differences are also reviewed.  Section~\ref{sec:groups} discusses the use of principal components in exploring the variation in surface data and in comparing groups.  Again a functional perspective is adopted, based on surface integration.  This section also discusses how effects can be visualised by characterising the shape changes associated with appropriate subspaces, rather than through examination of individual components.  Section~\ref{sec:individuals} addresses the situation where there is a need to assess the characteristics of individual surfaces, and in particular of any shape features which are not consistent with control shape.  Some final discussion and reflection is provided in Section~\ref{sec:discussion}.

The methods proposed in the paper are illustrated throughout on images of human facial shape.  There is a strong emphasis on the creation of visual displays which communicate patterns in the data, the evidence and nature of group differences, and the distinctive characteristics of individuals, as clearly as possible.  Graphics are provided in static form but animations are also available in the \textit{Supplementary Information}.


\section{Some fundamental tools for surfaces}
\label{sec:facial-model}

\subsection{Facial models}
\label{sec:facial-models}

A model for an individual surface should provide a structured representation of shape whose components have a consistent interpretation across the other surfaces in the dataset.  This then allows the investigation of pattern and variation in shape.  Landmarks satisfy this criterion and so, while the information they carry is limited, they have often been used as the starting point for more complex models.  \cite{paulsen2003shape}, \cite{hammond20043d} and \cite{mao-2006-patreclet} give examples of this approach where a template of a human face is `warped' onto an observed image.  Landmarks on the template are transformed in a non-linear manner to match those on the image exactly, with the surface of the template then adjusted further to improve the match with the surface of the image.  This might be done by locating closest points or by matching the characteristics of local surface curvature.  The resulting transformed template then provides a model for the surface whose meaning corresponds across all the images in the dataset.

\begin{figure}
\centerline{
   \includegraphics[width = 0.32\textwidth]{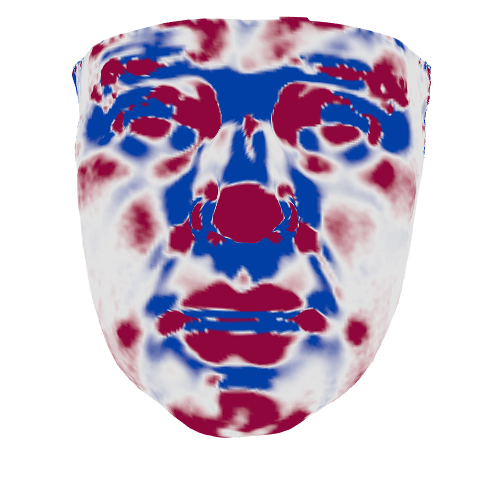}
   \includegraphics[width = 0.32\textwidth]{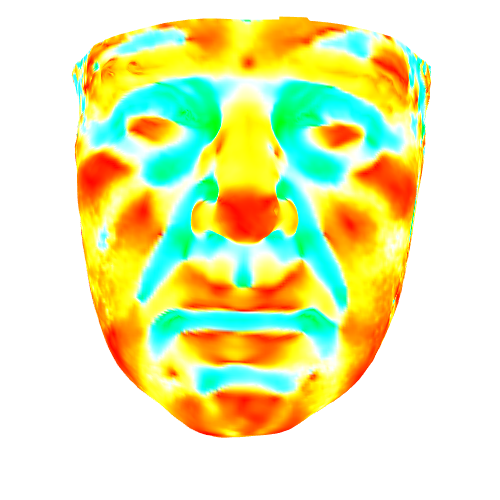}
   \includegraphics[width = 0.34\textwidth]{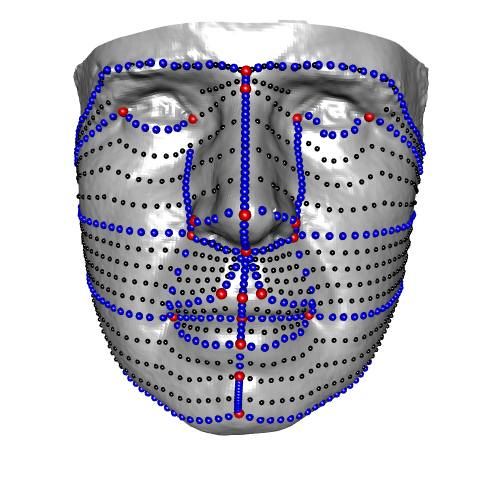}
}
\vspace*{-8mm}
   \includegraphics[angle = -90, width = 0.32\textwidth]{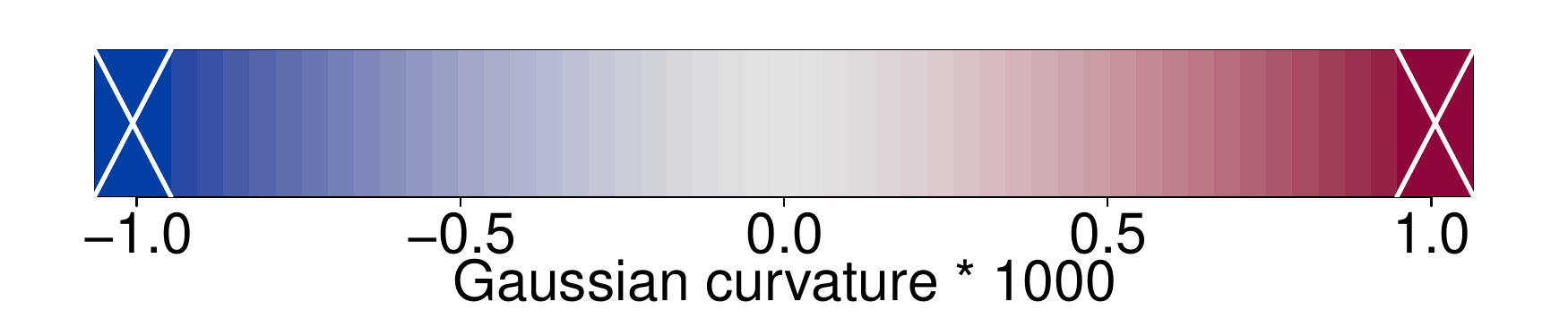}
   \includegraphics[angle = -90, width = 0.32\textwidth]{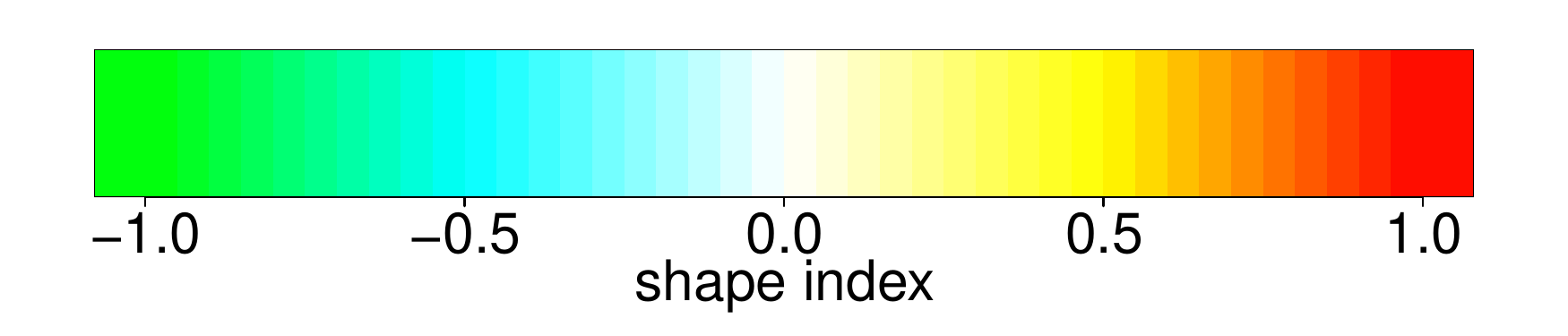}
\caption{The two left hand images show plots of Gaussian curvature and shape index.  The crosses at the end of the Gaussian curvature colour scale indicate that exceptionally high values have been truncated. The right hand image shows a fitted facial model, with landmarks (red), anatomical curves (blue) and surface patches (black) superimposed on an observed image (grey).}
\label{fig:curvatures}
\end{figure}

In an alternative approach, \citet{vittert-2019-aoas} took the view that ridge and valley curves provide the key information on shape, as these capture the locations where curvature is strongest.  The two left hand images in Figure~\ref{fig:curvatures} give examples of facial curvature, here in the form of Gaussian curvature and \textit{shape index}; see \citet{koenderink-1992-imagevisioncomputing} for details.  Curvature information can then be used to fit a model consisting of a set of ridge, valley or other geodesic curves, with landmarks as end-points.  A full surface representation can easily be constructed by interpolation across the surface patches bounded by these curves, although \citet{vittert-2019-aoas} give examples where the focused representation based on curves alone can be more informative.  An example of the resulting facial model is shown in the right hand panel of Figure~\ref{fig:curvatures}, which uses colour and size to indicate the hierarchical nature of the information captured in landmarks, curves and surface patches.

\subsection{Registration}

A key issue in the analysis of shape is that the observed images do not necessarily lie in a common co-ordinate system.  The process of data capture does not usually give each image the same origin or orientation.  The relative sizes of the images may also be viewed as unimportant from a shape perspective.  It is therefore necessary to remove these extraneous aspects before statistical analysis.  Different approaches to this are outlined below.  These are described in the context of transforming an image $X$ to match a reference image $Y$, where $X$ and $Y$ are $J \times 3$ matrices whose rows give the 3D positions of a fitted model in the discretised form of $J$ point locations.  The process of transforming $X$ to match $Y$ is referred to as registration.

\subsubsection{Procrustes methods}

A very effective approach is to find the rotation matrix $\Gamma$, scaling parameter $\beta$ and translation parameter $\gamma$ which bring $X$ as close to $Y$ as possible.  Adopting a similar notation to \citet{dryden-2016-book}, the method is expressed as:
\begin{equation}
      \min_{\beta, \Gamma, \gamma} || Y - \beta X \Gamma - 1_J \gamma^T ||^2 =
      \min_{\beta, \Gamma, \gamma} \sum_{j=1}^J ||y_j - \beta \Gamma^T x_j - \gamma ||^2,
\label{eq:procrustes}
\end{equation}
where $||.||$ is the Euclidean norm and $X$ and $Y$ are assumed to have centroid $0$.  This is a standard example of an approach referred to as Procrustes registration.  The ideas and methods of implementation involved are comprehensively described by \citet{dryden-2016-book}.  The solution to (\ref{eq:procrustes}) provides the basis of an iterative algorithm to match multiple images where $Y$ represents a mean shape and the criterion is summed over multiple images $X_i; i = 1, \ldots, n$.  A constraint on the size of $Y$ is adopted to avoid degenerate solutions.

These methods arose in the context of shape representations based on landmarks, often well-separated spatially.  The representations we are now dealing with may have a discrete point-based form, for convenience, but they represent a continuous surface.  This leads immediately to a functional data analysis perspective, as described by \citet{ramsay-1997-book}.  In the current setting, functional registration is achieved through
\begin{equation}
      \min_{\beta, \Gamma, \gamma} \int_{S_y} ||y -  \beta \Gamma^T x(y) - \gamma ||^2 dy,
\label{eq:opa-functional}
\end{equation}
where $S_y$ denotes the surface indexed by $Y$.  The function $x(y)$ indexes the point $x$ on the surface $S_x$ which has geometrical correspondence with the point $y$ on surface $S_y$.  The models for the two surfaces established this.  The integral can now be approximated in discrete form as
\begin{equation}
       \sum_{j=1}^J a_j ||y_j - \beta \Gamma^T x_j - \gamma||^2 = ||\sqrt{A} Y - \beta \sqrt{A} X \Gamma - \sqrt{A} 1_J \gamma^T||^2 ,
\label{eq:opa-weighted}
\end{equation}
where the weight $a_j$ gives the surface area which surrounds point $y_j$ and $A$ is a diagonal matrix containing the weights $a_j$.  These weights can be calculated easily.  If $\mathcal{T}_1, \ldots, \mathcal{T}_T$ denote the set of surface triangles, $\mathcal{N}_j$ is the set of indices of triangles which have $x_j$ as a vertex and $|.|$ denotes area, then the weights are simply $a_j = \frac{1}{3} \sum_{t \in \mathcal{N}_j} |\mathcal{T}_t|$.  The divisor $3$ apportions one third of the area of each triangle to each of its three vertices.

Expansion of the right hand side of expression (\ref{eq:opa-weighted}), following the derivation of the unweighted case in \citet{dryden-2016-book}, shows that the minimum is achieved when the matrices $AX$ and $AY$ are column-centred, with \begin{eqnarray*}
      \hat\gamma & = & 0, \\
      \hat\Gamma & = & UV^T, \\
      \hat\beta & = & \mbox{tr}\left\{Y^TAX\hat\Gamma\right\} / \mbox{tr}\left\{Y^TAX\right\} ,
\end{eqnarray*}
where $Y^TAX = ||\sqrt{A}Y|| ||\sqrt{A}X|| V \Lambda U^T$, with $\Lambda$ diagonal. \citet{dryden-2016-book} discuss more complex forms of weighting for other purposes.

The case of matching one shape $X$ to another $Y$ is referred to as \textit{ordinary Procrustes registration}.  This provides the building block for \textit{generalised Procrustes registration} which seeks a common registration of multiple shapes $X_1, \ldots, X_n$.  The aim now is to minimise the sum of the deviations of transformed shapes from a common mean $\mu$.  This can be expressed in functional form as
\begin{eqnarray*}
      & & \sum_{i=1}^n \int_{S_{\mu}} ||\mu - \beta_i \Gamma_i^T x_i(\mu) - \gamma_i ||^2 dy \\
      & & \phantom{xxxx}
      \approx \sum_{i=1}^n ||\sqrt{A} \mu - \beta_i \sqrt{A} X \Gamma_i - \sqrt{A} 1_J \gamma_i^T||^2,
\label{eq:opa-functional}
\end{eqnarray*}
with the transformation parameters $\beta_i, \Gamma_i, \gamma_i; i = 1, \ldots, n$.  The weights in the diagonal matrix $A$ are now the areas surrounding the vertices of the mean shape $\mu$.  Following again the general structure outlined in \citet{dryden-2016-book}, and beginning with each $AX_i$ column centred, minimisation can be achieved by successive ordinary weighted Procrustes registration of the adjusted shapes $X^P_i =  \beta_i X \Gamma_i - 1_k \gamma_i^T$ onto the mean $\mu$, which itself is estimated simply as the average of the $X^P_i$.  A size constraint is required to ensure that the solution does not degenerate to $0$.  From a functional perspective, size is expressed in the surface area of each shape and this is easily calculated in discrete form as the trace of $A$.  Notice that integration is carried out over the mean surface $\mu$, so that $A$ changes with each iteration.

\begin{figure}
\centering
\includegraphics[width = 0.24\textwidth]{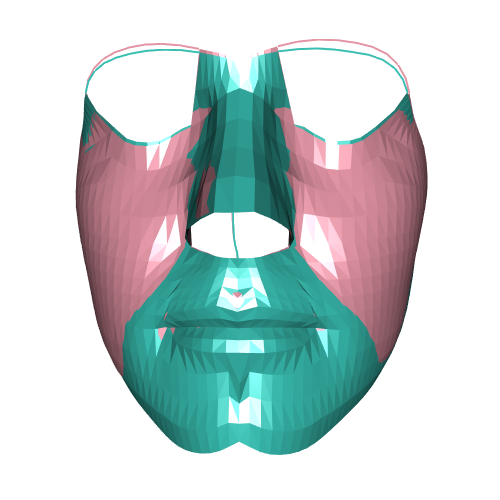}
\includegraphics[width = 0.24\textwidth]{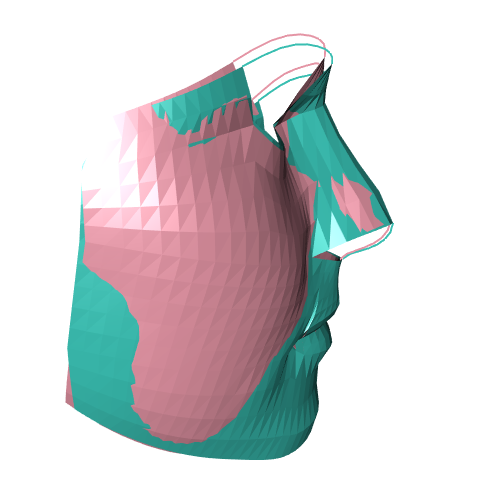}
\includegraphics[width = 0.24\textwidth]{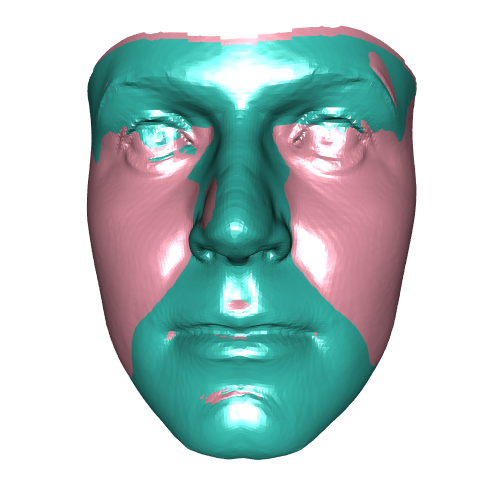}
\includegraphics[width = 0.24\textwidth]{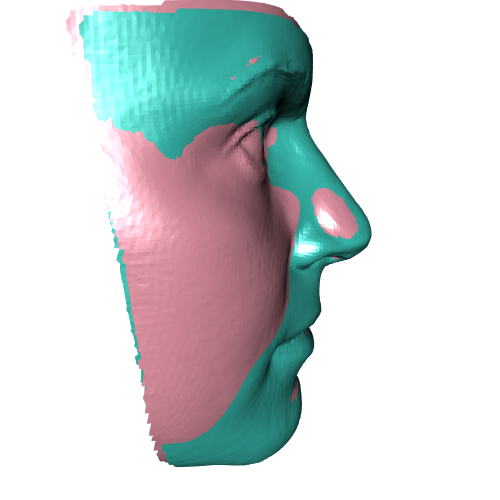}
\caption{Procrustes matched male (green) and female (pink) mean shapes, using model meshes (left, frontal and lateral) and warped facial templates matched to these mean shapes (right, frontal and lateral).}
\label{fig:sex-means}
\end{figure}

Figure~\ref{fig:sex-means} provides an illustration from a sample of $61$ males and $69$ females, all adults of UK origin, where Procrustes registration has been applied to each sex separately and the resulting means matched by a further Procrustes step.  Human sexual dimorphism has been extensively studied; see, for example, \cite{bruce1993sex}, \cite{wilkinson2004forensic}, \cite{Armann201269}, \cite{claes2012sexual}.  Figure~\ref{fig:sex-means} highlights the key differences on average, with males exhibiting greater prominence in nasal, chin and brow ridge areas while females correspondingly exhibit more prominent cheeks.

This simple example also raises a key issue of visualisation, namely how to compare two 3D surfaces most effectively.  The Figure adopts the simple strategy of superimposing the surfaces which gives a clear indication of which of the two shapes is more prominent in each area.  Other strategies will be considered in Section~\ref{sec:groups}.  Colour choice is also an important aspect of visualisation and this is discussed very helpfully by \citet{zeileis-2009-csda}, with effective solutions for comparing groups and displaying values on a continuous scale, including the presence of a reference value.  These colour choices have been adopted through the paper.

The facial models displayed in Figure~\ref{fig:sex-means} includes two curves at the brow and columella (between the nostrils), to avoid noisy areas of the image around the eyes and nostrils.  These curves are included in registration, and in later analysis, simply by considering a small patch around each curve point, with area set to the average of the areas surrounding the surface points.

\subsubsection{Warping}

Procrustes registration brings the co-ordinates of shape $X$ as close as possible to those of shape $Y$, within the limits imposed by the use of translation, rotation and scaling only.  However, there are situations where it is useful to match these two shapes exactly.  This arises in some of the methods involved in constructing facial models, described in Section~\ref{sec:facial-models}, where a template is initialised on  an observed image by exact matching of a set of landmarks.  A further example is in the improvement of the visual comparison of shapes such as those in Figure~\ref{fig:sex-means}.  If a high resolution facial template is available, with an embedded shape model $Z$ which corresponds to that of $X$ and $Y$, then a smooth function which transforms $Z$ to $X$ exactly can be identified, in a process known as \textit{warping}.  This function can then be applied to the template to creates a visual display which has smoother and more attractive surfaces than the lower resolution model $X$ and which adds in detailed features such as nostrils and eyes, giving a more effective and interpretable display of a human face.  While care should be taken not to interpret the form of these very detailed features, the principal characteristics of the display all reflect the underlying model.  Use of a template can also help to anonymise individual faces.  The right hand images in Figure~\ref{fig:sex-means} show the effects of employing templates in this way to both male and female means.  

In the analysis of 2D shapes based on landmarks, the concept of a deformation grid to describe shape change is a very old one; see \citet{darcy-1917-book}.  This uses a function which maps one set of landmarks to another exactly but, as it is expressed in functional form, this function can also be applied to a regular grid of locations over the first image to create a warped grid which expresses the underlying transformation.  Methods based on pairs of thin-plate splines were first introduced in 2D by \citet{bookstein-1989-ieeepami} and developed further by \citet{bookstein-1997-book}. The topic is also explained clearly by \citet{dryden-2016-book}.  Corresponding methods in 3D were first introduced by \citet{gunz-2005-edbook} and applied to skulls by \citet{mitteroecker-2004-jhumanevolution} and \citet{mitteroecker-2008-evolution}, to long bones by \citet{frelat-2012-amjphysanthrop}, and to mice heads by \citet{waddington-2017-cbnr}.  The literature on radial basis functions uses the same techniques but employs a different language.  The technical details of warping in 3D are described in the Appendix.

\subsection{Visual comparison of two shapes}

\begin{figure}
\centerline{
   \includegraphics[width = 0.22\textwidth]{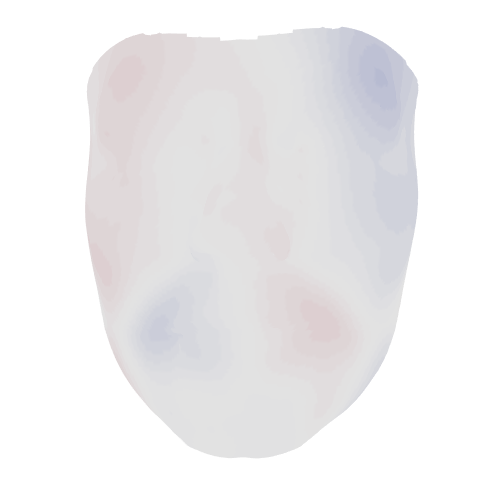} \hspace{-6mm}
   \includegraphics[width = 0.22\textwidth]{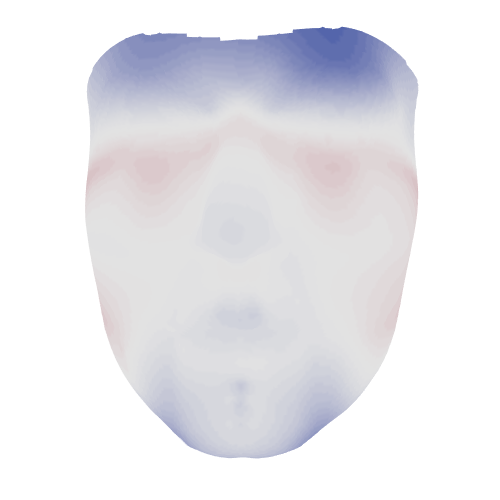} \hspace{-6mm}
   \includegraphics[width = 0.22\textwidth]{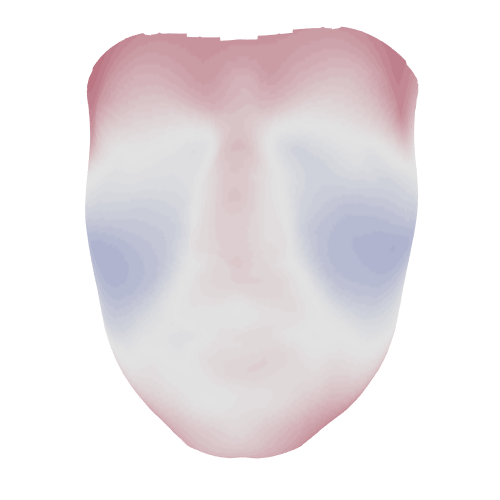} \hspace{-6mm}
   \includegraphics[width = 0.22\textwidth]{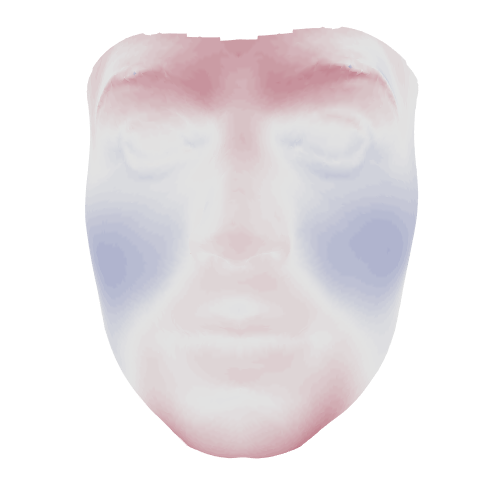} \hspace{-6mm}
   \includegraphics[width = 0.22\textwidth]{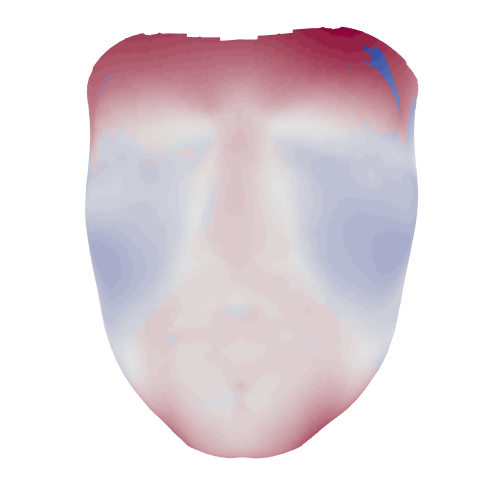} \hspace{-6mm}
   \includegraphics[width = 0.045\textwidth]{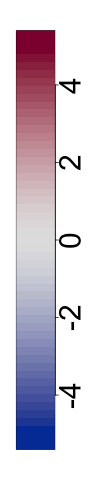}
   }
\textsf{\hspace{1.1cm} x \hspace{2.2cm} y \hspace{2.2cm} z \hspace{1.8cm} normal \hspace{0.4cm} signed~Euclidean}
\caption{The mean female face painted with colours to represent the movement (mm) required in different dimensions to reach the mean male face.}
\label{fig:sex-means-colour}
\end{figure}

The male and female example of Figure~\ref{fig:sex-means} raises the question of how two shapes can most effectively be compared visually.  The challenge is that in addition to the 3D shapes themselves, comparison involves an additional vector field of differences, with a displacement vector at each position on the individual shapes.  A helpful strategy is to display one shape and use colour to inform on the shape difference from the other shape at each location.  The lower rows of Figure~\ref{fig:sex-means-colour} illustrate this by plotting the female mean face and using colour to indicate distances to the male mean.  The distances in the separately $x$, $y$ and $z$ co-ordinates are shown.  These co-ordinates can be given clear interpretations by orienting the female mean so that nominated landmarks such as the outer corners of the eyes (\textit{exocanthions}) define the direction of the $x$-axis and others such as the top of the nose ridge between the eyes and the central point at the base of the nose (\textit{nasion} and \textit{subnasale}) define the $y$-axis.  The final two images use colour to indicate the distance between the corresponding points on the female and male means projected along the normal direction at the surface of the female mean, and the Euclidean distance between corresponding points, with sign determined by whether the projection along the normal is positive or negative.

None of these devices captures the information in the shape difference completely, because the change is in 3D while the colour scale can represent only a single dimension, but they provide options for detailed exploration.  The simple superimposition of surfaces illustrated in Figure~\ref{fig:sex-means} is usually a good place to start as it shows the broad, qualitative differences between the shapes, with the other options available as follow-up.  For small scale movements, the normal and signed-Euclidean distances are often effective and can give greater detail on the nature and size of the movement, as the two surfaces are generally close.  Other plotting devices are available, such as the use of transparent surfaces, or one transparent surface with a wireframe representation.  The best choice of display will depend on the particular features and differences of the shapes involved.

One of the most effective means of displaying differences is through animation, with the display of a sequence of intermediate steps along a path between the two shapes to be compared.  Several of the plots in this paper have animated versions which are available in the \textit{Supplementary information}.


\section{Visualising shape datasets}
\label{sec:groups}

\subsection{Exploring variation}

\begin{figure}
	\centerline{
		\includegraphics[width = 0.3\textwidth]{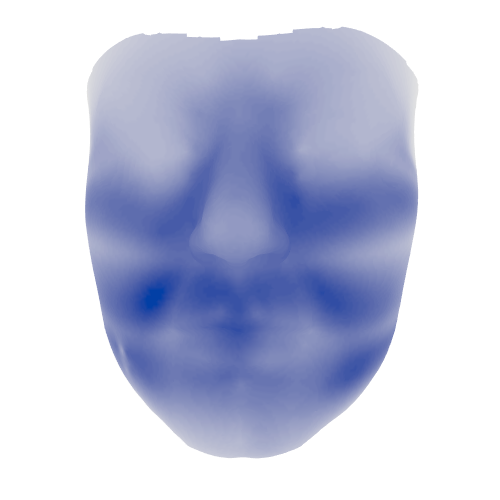}
		\includegraphics[width = 0.064\textwidth]{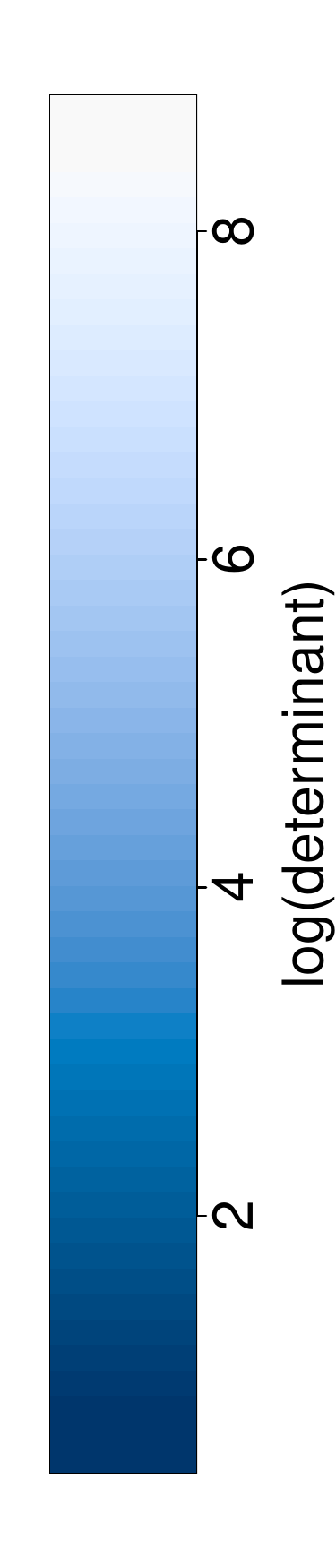}
	}
	\caption{The variability in British female faces.}
	\label{fig:variability}
\end{figure}

While a visual comparison of means is useful, an understanding of the variation involved in a dataset is necessary for any form of statistical analysis.  A simple device is to display the size of the variation at each location on the model.  Figure~\ref{fig:variability} shows the value of $\log |\det(\hat\Sigma_j)|$, where $\hat\Sigma_j$ is the empirical covariance matrix of the $x$-, $y$- and $z$-coordinates at location $j$ after Procrustes registration.  The regions of higher variability include the eyes, whose reflective surface can introduce some inaccuracy, the forehead, which lies at the edge of the facial surface, and the chin and nasal tip, where the degree of prominence can vary considerably.  Effects associated with the model curves which traverse the cheeks, where flatness can induce some variability in location, are also apparent.

Descriptions which capture the correlation between locations are clearly required.  These also need to deal with the difficulty that the dimensionality of the shape representation (for example, $917$ 3D points in a discrete representation of a surface) often exceeds by a large margin the number of shapes present.  Principal components offers an immediate route to the creation of a lower dimensional space which captures the principal features of shape variation.  Procrustes registration places the aligned shapes $X_i$ in a non-linear space but the \textit{approximate tangent co-ordinates}, $\mbox{vec}(X_i - \bar{X})$, can be analysed very effectively as a linear space.  Here the $\mbox{vec}$ operator creates a vector of length $3J$ from the $3 \times J$ matrix $X$ by stacking its columns.  The eigenvectors $e_l$ and eigenvalues $\lambda_l$ of the covariance matrix of the tangent co-ordinates then capture the directions along which variation is sequentially maximal.  For the $k$th direction, a description of the variation involved is provided by considering $c \sqrt{\lambda_k} e_k$ which, when reassembled into a $J \times 3$ matrix using the $\mbox{vec}^{-1}$ operator, represents shape variation from the mean along the principal component direction in multiples of the standard deviation.  The multiplier $c$ is often set at $\pm 2$, or $\pm 3$ if some magnification is required.  \citet{dryden-2016-book} give all the details.

A functional data analysis perspective can be applied in this setting by following the pattern described by \citet{ramsay-1997-book}.  When the data are in the form of functional objects, $x(s)$, where $s$ lies in an appropriate sample space $\mathcal{S}$, principal components are then defined as orthonormal functions $\beta(s)$ which successively maximise the variance of $\int \beta(s) x(s) ds$.  In many settings, the sample space $\mathcal{S}$ is a time interval or a spatial region in standard Cartesian form.  In the present setting, the functional object has the much more complex form of a 2D manifold embedded in 3D space.  The immediate problem is how to parametrise this in a suitable sample space $\mathcal{S}$.  A solution is provided by setting this to be the Procrustes mean shape, $\mathcal{S}_{\mu}$.  Any other shape in the sample can then be expressed through the three functions $\{x(s), y(s), z(s)\}$ which give the 3D deviations of this shape from the mean at location $s$.

This takes us to the realm of multivariate functional principal components which seek to maximise the variance of 
$$
           \int_{\mathcal{S}_{\mu}} \beta_x(s) x(s) ds + \int_{\mathcal{S}_{\mu}} \beta_y(s) y(s) ds + \int_{\mathcal{S}_{\mu}} \beta_z(s) y(s) ds ,
$$
as discussed by \citet{ramsay-1997-book}.  As usual, computations are conveniently based on discrete approximations to these integrals.  The model form of each shape has a consistent triangulation so, for example, a convenient approximation can be written as
$$
 		\int_{\mathcal{S}_{\mu}} \beta_x(s) x(s) ds \approx \sum_j \beta_x(s_j) x(s_j) a_j ,
$$
where $a_j$ is the area surrounding $s_j$.

\begin{figure}
\begin{minipage}{0.6\textwidth}
   \includegraphics[width = \textwidth]{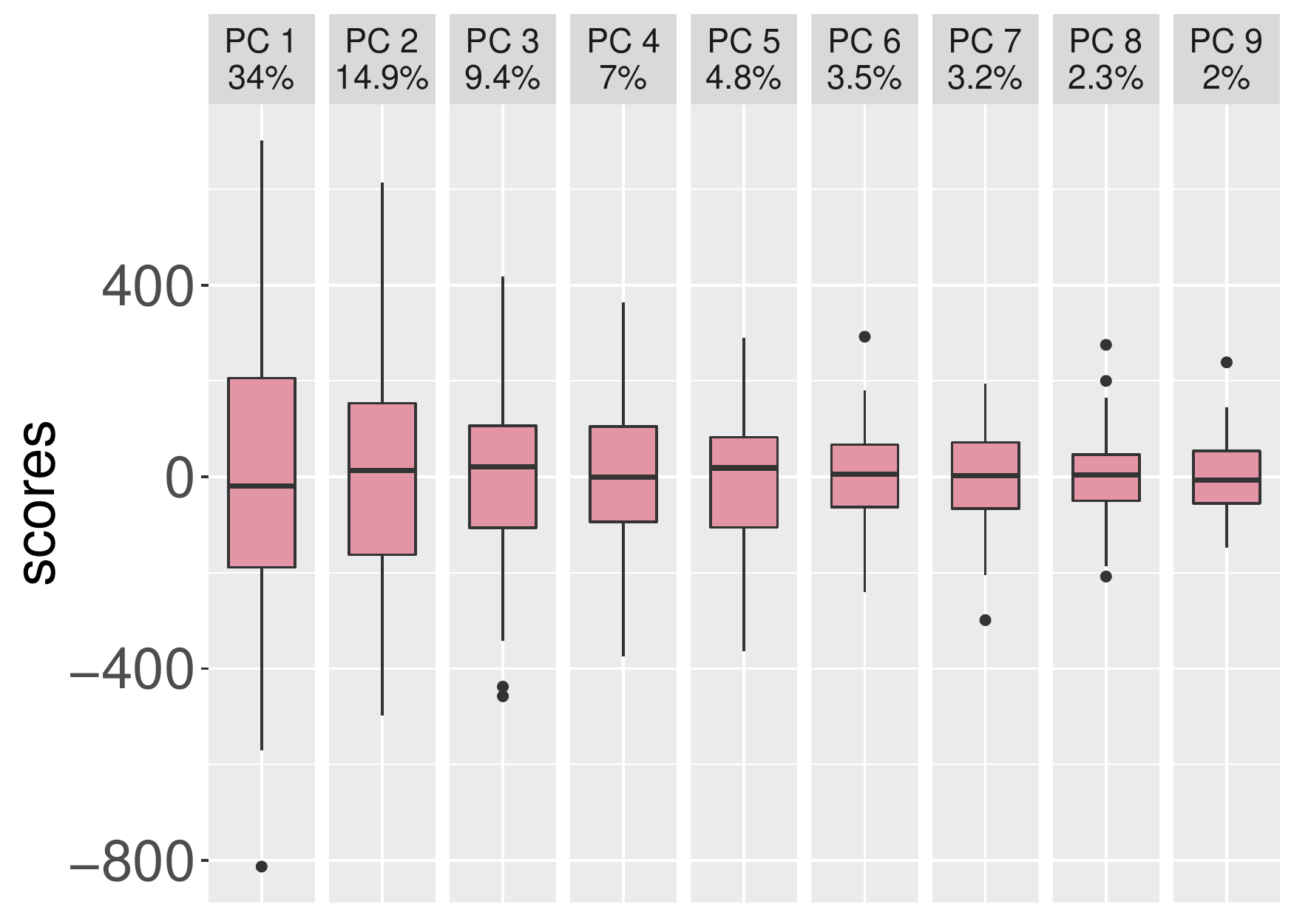}
\end{minipage}
\begin{minipage}{0.4\textwidth}
   \centerline{\textsf{\small PC 1 \hspace{1.6cm} PC 2}}
   \centerline{
      \includegraphics[width = 0.45\textwidth]{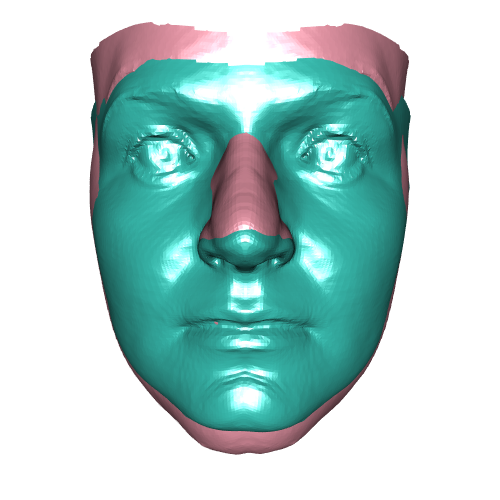}
      \includegraphics[width = 0.45\textwidth]{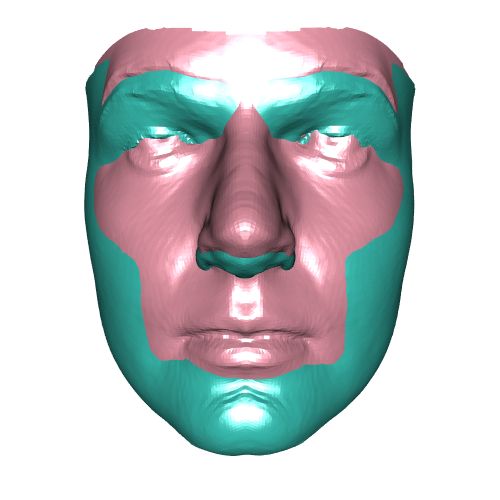}
   }
   \centerline{\textsf{\small PC 3 \hspace{1.6cm} PC 4}}
   \centerline{
      \includegraphics[width = 0.45\textwidth]{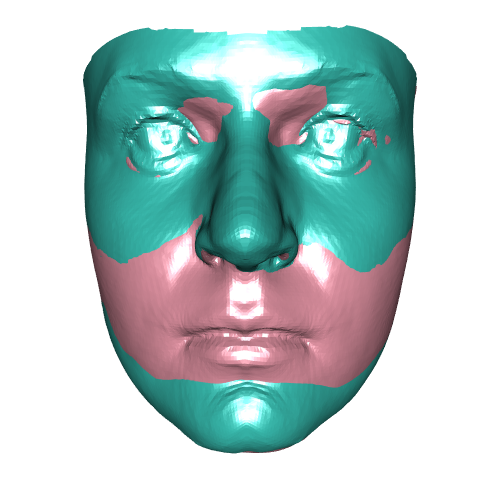}
      \includegraphics[width = 0.45\textwidth]{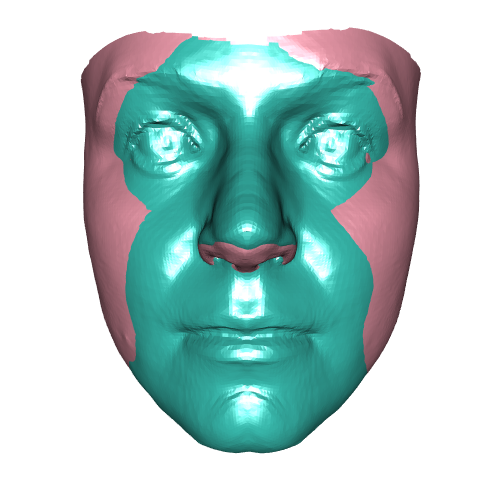}
   }
\end{minipage}
\centerline{
   \includegraphics[width = \textwidth]{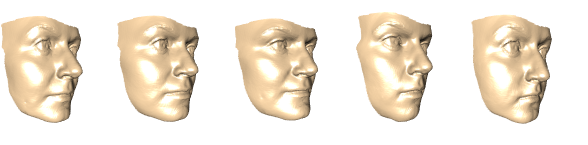}
}
\caption{The top left hand plot shows the scores for the first $10$ principal components for British females.  The four top right hand plots show the nature of the shape change associated with the first $4$ principal components.    The lower plots show randomly generated faces from a `grand tour' of the variation in British female faces.  See the \textit{Supplementary information} for animations.}
\label{fig:grand-tour}
\end{figure}

Figure~\ref{fig:grand-tour} shows the results of applying functional principal components to the British female data.  Given the high dimensionality of shape surfaces, the number of components required to capture a high proportion of the variation in the data may be reasonably large, with $10$ components required to capture $82$\% of the variation in this case.  Figure~\ref{fig:grand-tour} shows the scores, $\mbox{vec}(X_i - \bar{X})^T e_k$, on these principal components, with the diminishing widths of the boxplots illustrating the gradual reduction in variation across the components.  The shape changes associated with the first four principal components are indicated by superimposing the faces which correspond to $c=\pm 2$ standard deviations (pink and green).  Methods for investigating individual components are discussed below in the context of comparing groups but the variation in a single group can be helpfully displayed through the idea of a `grand tour', proposed for general multivariate data by \citet{asimov-1985-siam}.  A very simple version of this uses a vector of $p$ independent normal random variables $z$ to create a random sample of locations in the space of the first $p$ principal components, $\{z_k \sqrt{\lambda_j} e_k; k = 1, \ldots, p\}$.  Turning these into shapes and tracking between successive pairs by simple interpolation creates an animation which randomly explores the variation in shape.  Figure~\ref{fig:grand-tour} illustrates five random positions which form the staging posts of a tour.  This approach forms the basis of a comparison between individuals and a control dataset in Section~\ref{sec:individuals}.

\subsection{Assessing differences between groups}

\begin{figure}
	\includegraphics[width = 0.5\textwidth]{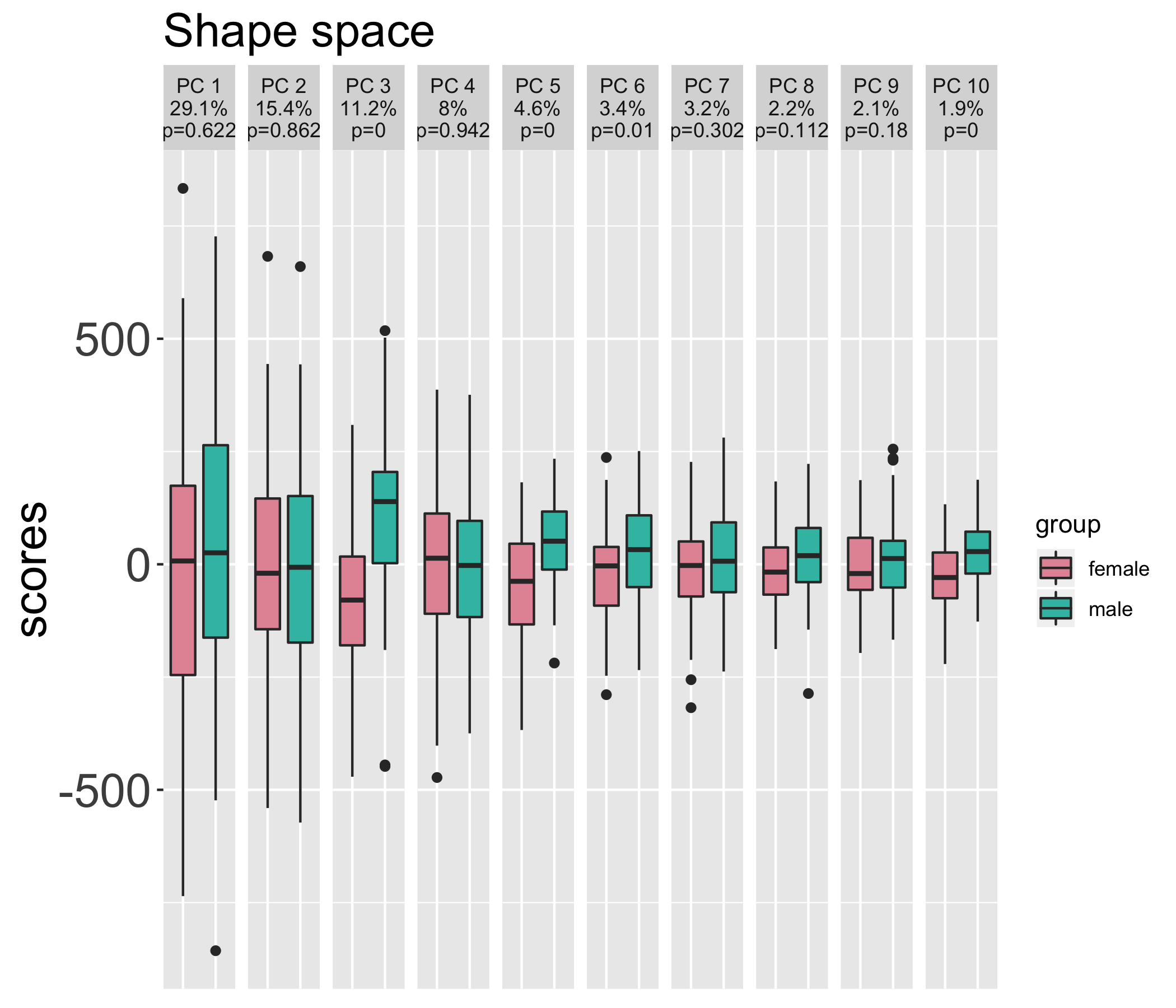}
	\includegraphics[width = 0.5\textwidth]{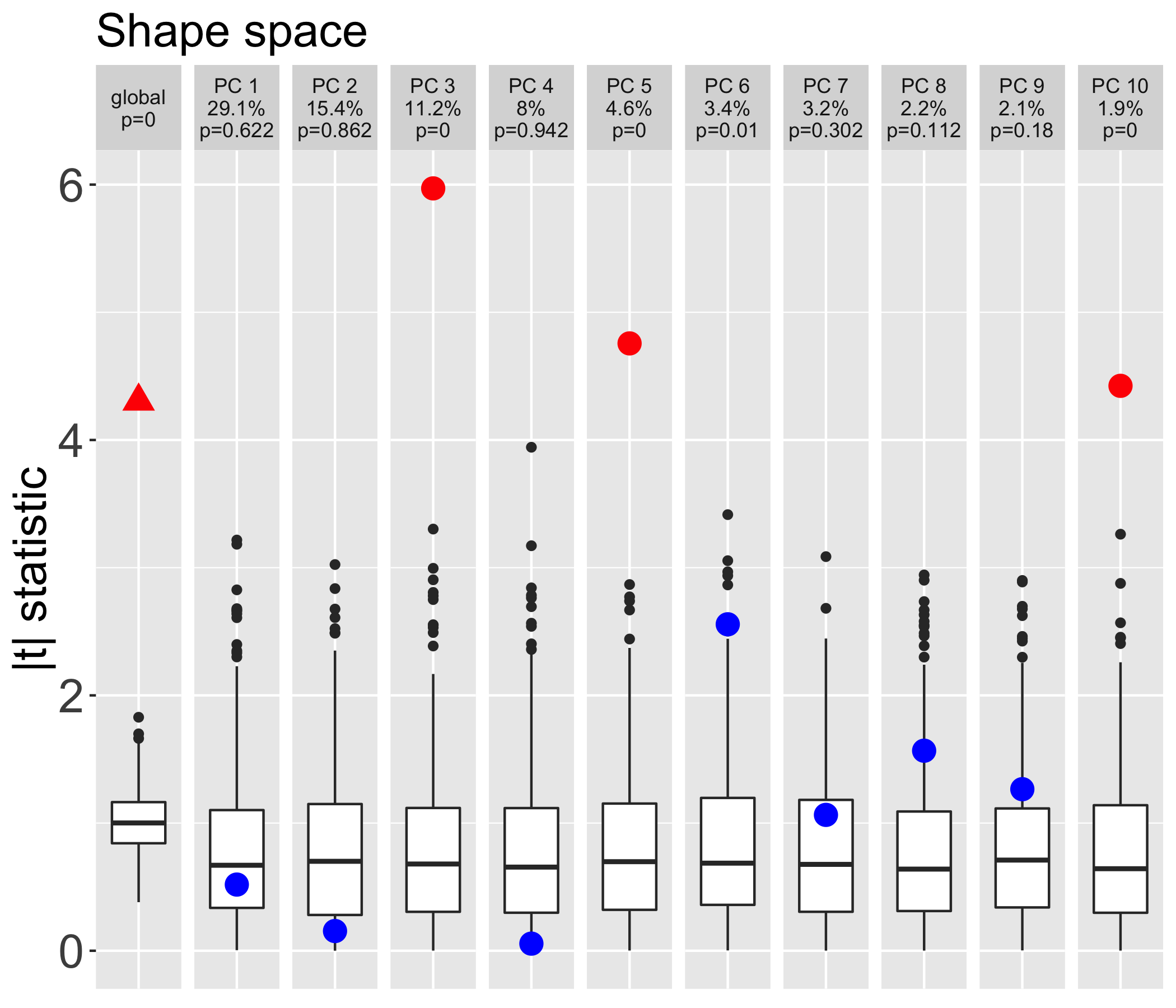}
	
	{\tiny \hspace{2.2cm} \textsf{PC 3} \hspace{2.3cm} \textsf{PC 5} \hspace{2.2cm} \textsf{PC 10}
           \hspace{2.0cm} \textsf{Combined} } \\
      \centerline{
		\includegraphics[width = 0.2\textwidth]{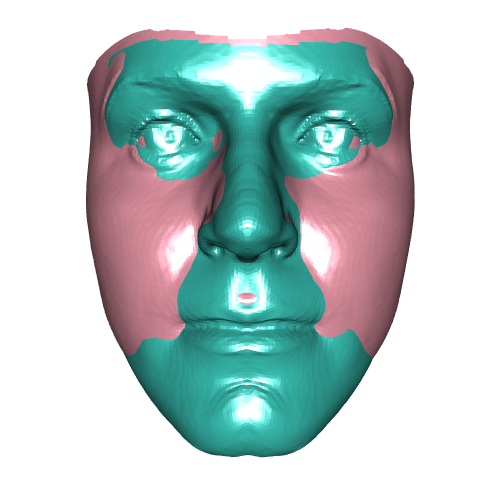}
		\includegraphics[width = 0.2\textwidth]{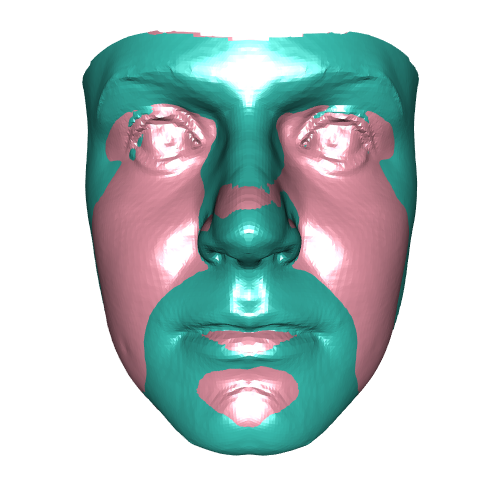}
		\includegraphics[width = 0.2\textwidth]{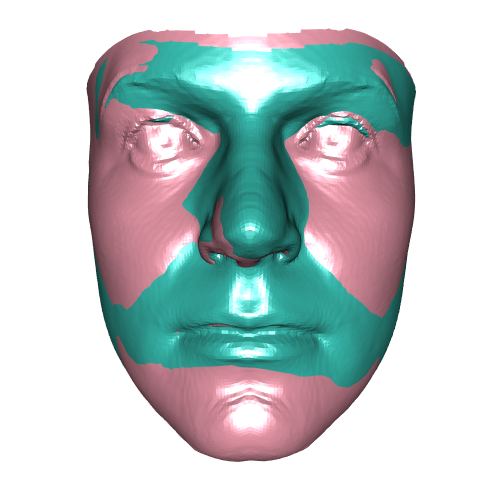}
		\hspace{1mm}
		\includegraphics[width = 0.2\textwidth]{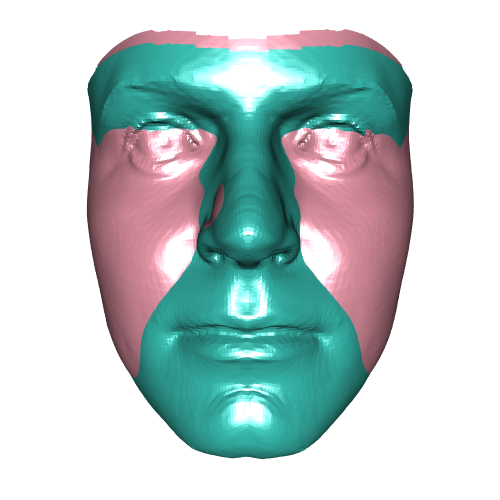}
	}
\caption{The top row shows the scores for males and females for the first $10$ principal components constructed in tangent space and the simulated test statistics (boxplots) and observed values (triangle and dots) for a test of identical distributions.  Test statistics showing evidence for differences are plotted in red. The second row shows the shape change associated with each significant component and their combination.  Green and pink refer to the male and female ends of the scale respectively.}
\label{fig:sex-shape}
\end{figure}

When groups representing different populations are present in a dataset, principal components provide a helpful way of reducing the dimensionality of the space in which comparison takes place, while retaining as much of the variability as possible.  If components are simply constructed from the combined dataset, without reference to the group structure, then the variation captured by each component will contain both intra- and inter-group contributions.  The top left hand plot in Figure~\ref{fig:sex-shape} displays the scores for the principal components constructed in this way for the sexual dimorphism data.  As the signs of the eigenvectors which define the components are arbitrary, these have been reversed where necessary to ensure that the male mean score is higher, for ease of interpretation.  It is often the case that the first few components capture large scale variation (greater width and smaller height etc.) which is common across groups, with group differences associated with more subtle aspects of shape.

In the reduced space of the first $p$ components, a global assessment of the evidence for mean differences in male and female shape is provided by Hotelling's $T^2$ statistic, $T^2 = \frac{1}{(1/n_m + 1/n_f)} \left(\bar{v}_m - \bar{v}_f \right)^T \hat\Sigma^{-1} \left(\bar{v}_m - \bar{v}_f \right)$, where $\bar{v}_m$, $\bar{v}_f$ denote the mean $p$-dimensional score vectors, $n_m$, $n_f$ denote the group sample sizes and $\hat\Sigma$ denotes the usual estimate of the common covariance matrix of the scores.  It is also tempting to explore the nature of any evidence of differences by examining the $t$-statistics, $\frac{\left(\bar{v}_{lm} - \bar{v}_{lf} \right)}{(1/n_m + 1/n_f) \hat{\sigma}_l}$, where $\hat{\sigma}_l$ denotes the estimate of the common standard deviation of the groups on the $l$th component.  However, \citet{bedrick-2019-biometrics} demonstrates that care needs to be exercised because the distributional properties of test statistics are affected by the construction of the component directions in terms of optimising variance.  In light of this, a permutation approach is attractive.  Here the reference distributions for the test statistics are generated empirically simply by recomputing the values from the dataset with group labels randomly permuted $500$ times.  The top right hand panel of Figure~\ref{fig:sex-shape} illustrates this.  In order to make the distribution for the global test comparable with the others, the values of $\sqrt{T^2/p}$ have been plotted and the individual $t$-statistics have been placed on an absolute value scale.  The values of the test statistics computed from the observed data are shown as a triangle (global) and dots (components), with the empirical p-values noted at the top of the plot.  The global test gives strong evidence of differences in means while components $3$, $5$ and $10$ are identified as the strongest sources of difference.  To adjust for the multiple comparisons involved, a Bonferroni threshold for the empirical p-values has been adopted as $0.05 / 10 = 0.005$ and colour (red) has been used to indicate where that threshold has been exceeded.    As \citet{bedrick-2019-biometrics} points out, it is important to note that the p-values associated with individual components should be interpreted in the context of the null hypothesis that the mean scores are identical for \textit{all} components simultaneously.

The images in the lower part of Figure~\ref{fig:sex-shape} indicate, for those components which exhibit evidence of differences in groups, the nature of the associated shape change in the usual form of $\pm2 \sqrt{\lambda_k}$ from the mean.  The image corresponding to the positive end of the scale is more strongly associated with male shape (green) and the negative end of the scale with female shape (pink).  The association of male shape with more prominent nose, chin and eyebrows, and female shape with more prominent cheeks, is apparent.  However, the individual components cannot be given special status in the description of male-female differences as they were constructed simply by maximising the variation explained across the whole dataset.  It is therefore helpful to construct a combined display which corresponds to movement along all these components simultaneously.  This is aided by the earlier modification of components to ensure that positive signs are more strongly associated with males.  Movement to $\pm2 \sqrt{\lambda_k}$ in all components simultaneously would construct a rather extreme shape so the values $\pm2 \sqrt{\lambda}_k / \sqrt{q}$ are used, where $q$ denotes the number of components in simultaneous movement (here $q=3$).  This ensures that the resulting shape sits on the same quantile contour of a multivariate normal distribution as the shapes which move the individual components to $\pm 2 \sqrt{\lambda_k}$.  The result for the sexual dimorphism data is shown at the right hand end of the row of facial images in Figure~\ref{fig:sex-shape}.  This gives a very helpful representation of the combined effects of the individual components which carry evidence of differences.  It also has the attractive property of giving stronger weight to those components which explain larger amounts of variation.  The overall difference in shape change is clear and corresponds closely to the comparison of means in Figure~\ref{fig:sex-means} but this is now backed up by convincing statistical evidence.

When group differences are of interest, an alternative approach to principal components is through the intra-group covariance.  As pointed out by \citet{dryden-2016-book}, the $T^2$ statistic can be written as
$$
      T^2 = \sum_{k=1}^p \left[ \frac{\bar{v}_{k1} - \bar{v}_{k2}}{\sqrt{\lambda_k  (1/n_1 + 1/n_2)}} \right]^2 ,
$$
where the $\lambda_k$ denote the eigenvalues and $\bar{v}_{k1}$, $\bar{v}_{k2}$ the mean principal component scores, using the eigenvectors derived from an estimate of the common covariance matrix.  Dimensionality reduction follows from the truncation to $p$ terms, with each individual term having the attractive interpretation of the square of a two-sample $t$-statistic on the scores from each component.  The warning of \citet{bedrick-2019-biometrics} about distributional properties again applies, with a permutation test providing a convenient solution.  However, this time the eigen-decomposition needs to be performed for every random permutation because estimation of the common covariance matrix depends on the group structure.  

\begin{figure}
	\centering
	\begin{minipage}{0.4\textwidth}
		\includegraphics[width = \textwidth]{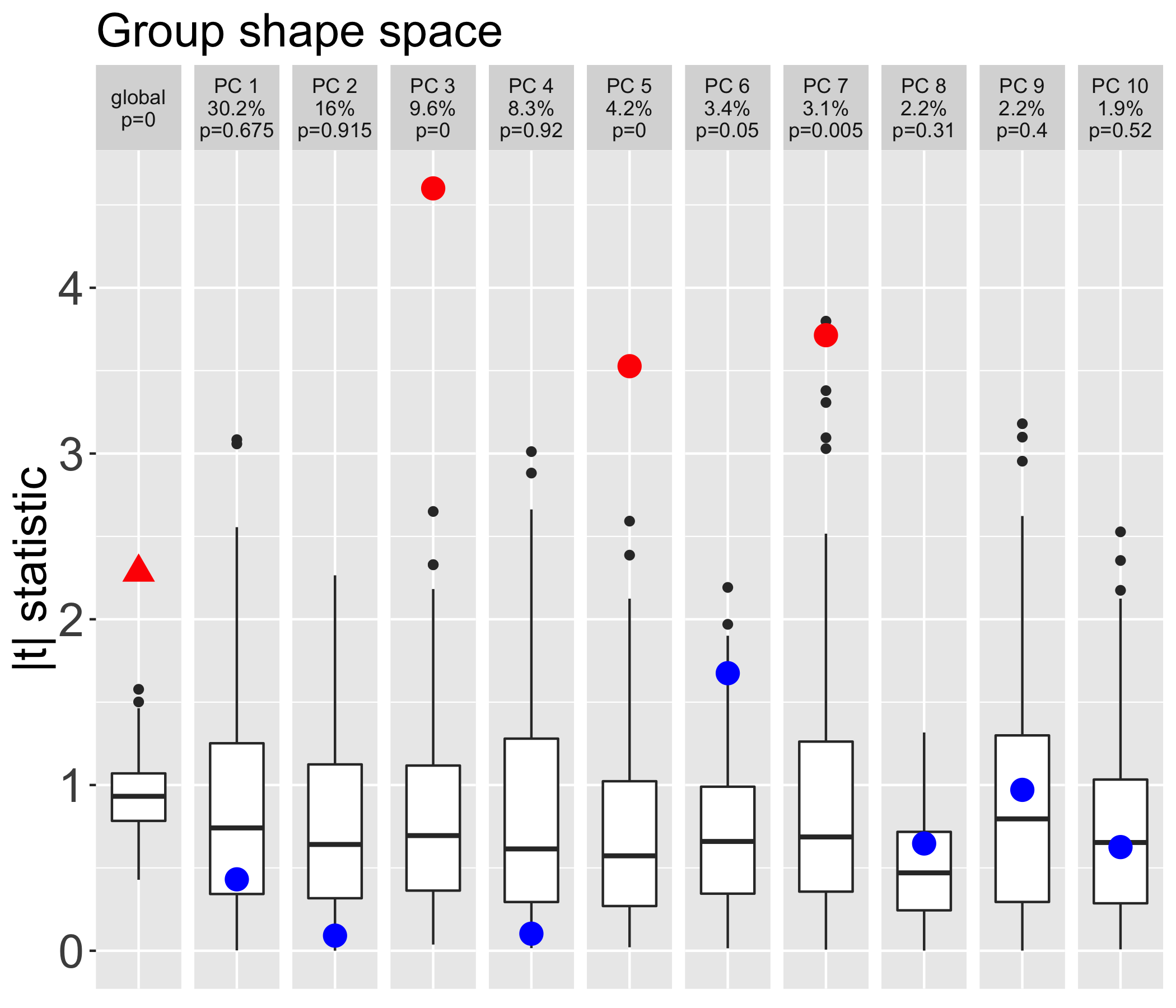}
	\end{minipage}
	\begin{minipage}{0.25\textwidth}
		{\tiny \hspace{0.5cm} \textsf{PC 3} \hspace{1.1cm} \textsf{PC 5}} \\ 
		\includegraphics[width = 0.48\textwidth]{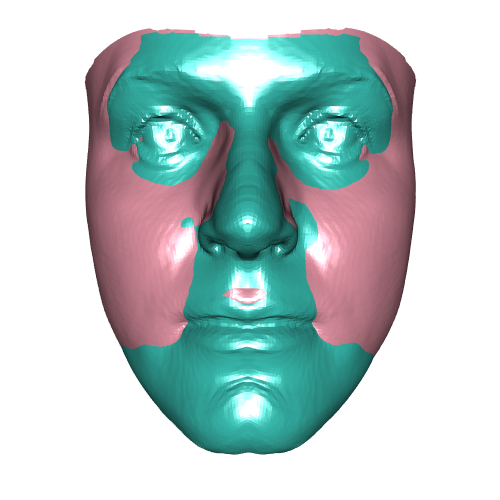}
		\includegraphics[width = 0.48\textwidth]{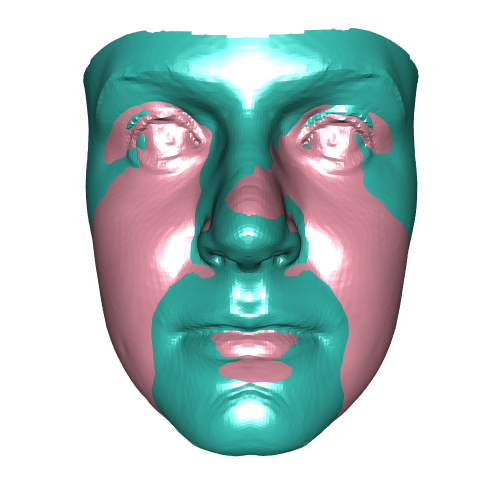} \\
		{\tiny \hspace*{0.5cm} \textsf{PC 7}} \\
		\includegraphics[width = 0.48\textwidth]{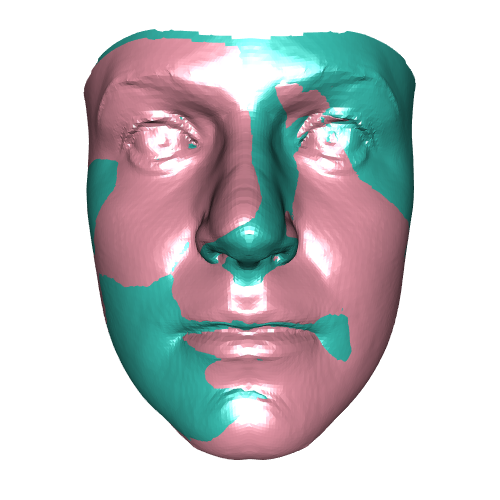}
	\end{minipage}
	\begin{minipage}{0.3\textwidth}
 	{\tiny \hspace{1.5cm} \textsf{Combined}} \\
	\includegraphics[width = \textwidth]{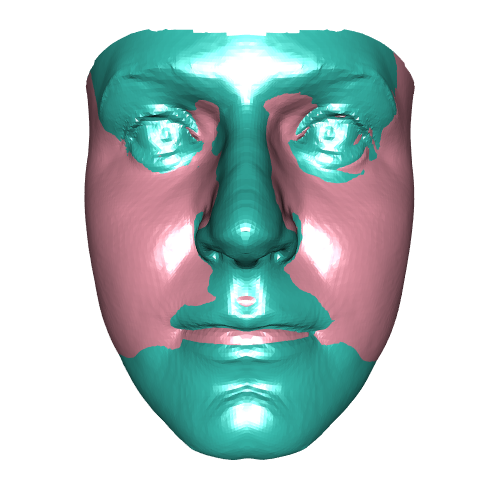}
	\end{minipage}
\caption{The left hand plot shows the simulated test statistics (boxplots) and observed values (triangle and dots) for a test of identical male and female distributions, using the components derived from the common covariance matrix.  The small faces shows the shape change associated with each significant component and the larger face illustrates their combined effect.}
\label{fig:sex-group-shape}
\end{figure}

Figure~\ref{fig:sex-group-shape} shows the results of this `group shape space' approach on the sexual dimorphism data, with $p=10$ to remain consistent with the earlier example.  The absolute value scale has been used again for the boxplots.  This loses the property that the global statistic is a simple average of its components but the $t$-statistic scale is helpful, and there is no effect on the performance of the tests.  The smaller facial images show the nature of the shape change associated with the individual components ($3$, $5$, $7$) where there is strong evidence of differences between males and females.  There is no reason why the differences in mean shape should align with the axes of the common covariance matrix so, again, the individual components do not have special status.  The larger facial image shows the shape change associated with the combination of these three components.  This characterises the sub-space where the evidence for difference is strongest and it is reassuring to see that this is very similar to the sub-space identified from the principal components which do not exploit group structure, as displayed in Figure~\ref{fig:sex-shape}.  This underlines the case for identifying and interpreting the sub-space as a whole, with the components simply providing particular indexing bases.

\subsection{Affine/non-affine decomposition}

A further sub-space approach is available through partitioning the variation in the data into affine and non-affine components.  The former involves linear transformations which apply across the whole object of interest.  The latter space contains non-linear transformations which describe local and more complex effects.  \cite{rohlf2003computing} showed that these sub-spaces can be easily created from the Procrustes aligned shapes $\{X_i; i = 1, \ldots, n\}$ through the regression models
$$
      X_i = \bar{X} \alpha_i + \varepsilon_i ,
$$
where the $\alpha_i$ denote $3 \times 3$ matrices of regression coefficients.  The affine co-ordinates $X_{Ai}$ are then available as the fitted values while the non-affine co-ordinates $X_{Ni}$ are obtained by adding the residuals to the mean as
\begin{eqnarray*}
      X_{Ai} & = & \bar{X} \hat{\alpha}_i , \\
      X_{Ni} & = & \bar{X} + (X_i - X_{Ai}) ,
\end{eqnarray*}
where $\hat{\alpha}_i$ denotes the least squares estimates.  More formally, the algebra associated with linear regression, particularly the independence of residuals and fitted values, separates the space of the Procrustes shape co-ordinates $X_i$ into two orthogonal sub-spaces which capture the affine and non-affine behaviours.  Analysis can therefore proceed separately within these sub-spaces to provide complementary descriptions of the variation in the dataset.

\begin{figure}
{\tiny \hspace{0.6cm} \textsf{Chinese} \hspace{2.1cm} \textsf{British} \hspace{2.1cm} \textsf{tangent} \hspace{2.1cm} \textsf{affine} \hspace{2.0cm} \textsf{non-affine}} \\ 
\centerline{
	\includegraphics[width = 0.2\textwidth]{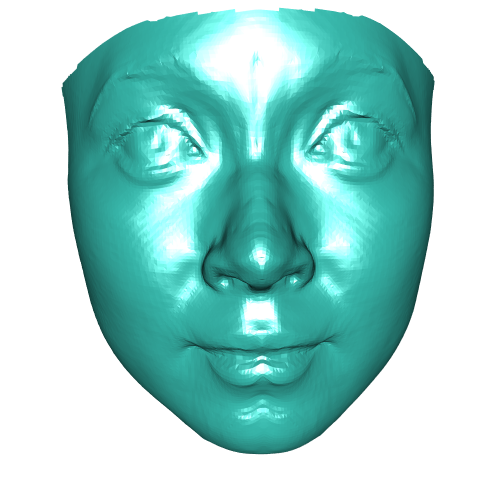}
	\includegraphics[width = 0.2\textwidth]{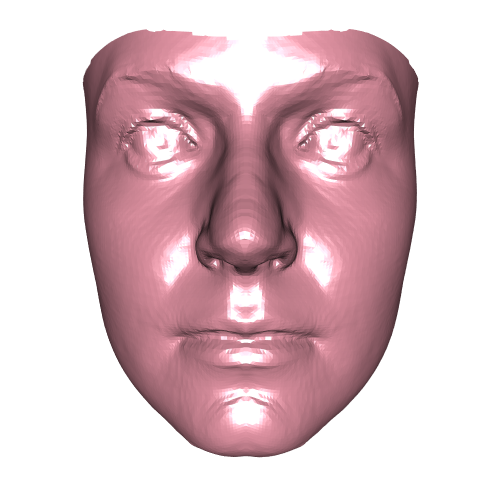}
	\includegraphics[width = 0.2\textwidth]{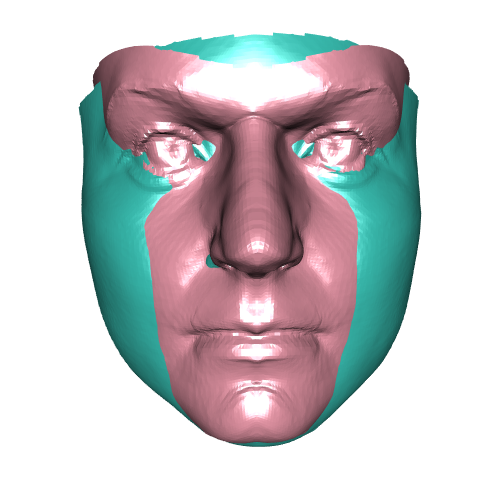}
	\includegraphics[width = 0.2\textwidth]{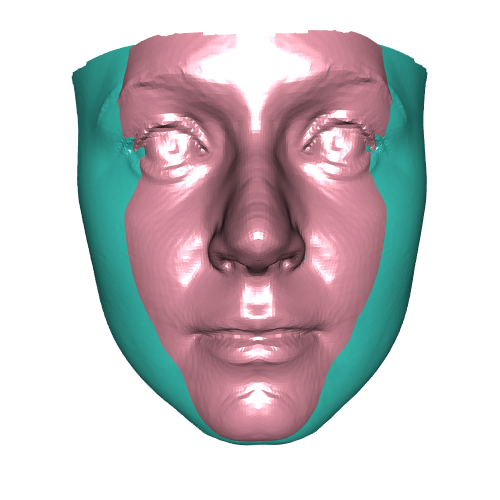}
	\includegraphics[width = 0.2\textwidth]{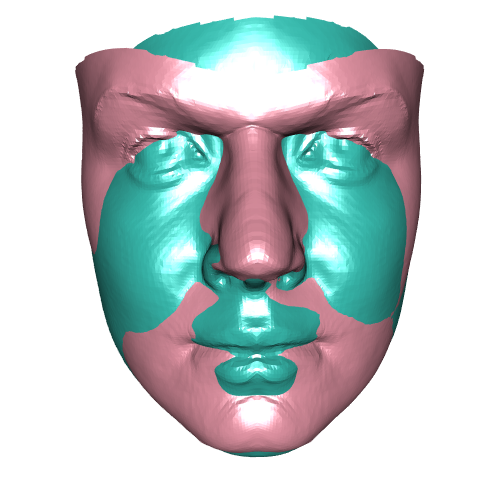}
}
\caption{From left to right, the images show the mean Chinese female face, the mean British female face, and the combined principal components of shape change in tangent, affine and non-affine spaces respectively.}
\label{fig:chinese-british}
\end{figure}

The comparison of British and Chinese female facial shapes provides a simple example.  Visual discrimination between these two ethnic groups is usually straightforward but examination of mean shapes allows the distinctive features to be identified and quantified.  Figure~\ref{fig:chinese-british} shows the nature of shape change in different sub-spaces, based on a sample of $69$ British and $26$ Chinese adult subjects.  In overall tangent space, there is only one significant component, with the associated shape change displayed in the central image in the usual form of $\pm 2$ standard deviations around the mean.  This draws attention to the more prominent central areas in British subjects (pink), including the brow ridge, nose, oral region and chin.  Correspondingly, the overall shape of Chinese faces (green) is flatter than that of their British counterparts, with more prominent cheek areas.  This is an example where inter-group differences dominates the variation in the data.  Indeed, the strength of this difference is indicated by the fact that there is no overlap between the scores of the British and Chinese faces on the first principal component in tangent space.

It is interesting to explore whether these differences can be explained by affine transformation or whether non-affine transformations are required.  In the affine sub-space only the first principal component shows clear evidence of difference between the groups and it is already clear that the lower brow of the British faces is not captured in this sub-space.  This is confirmed by analysis in the non-affine sub-space where there are two principal components which exhibit clear evidence of differences between the groups and whose combined effects are displayed in the right hand image of Figure~\ref{fig:chinese-british}.

\section{Visualising the shape of individuals}
\label{sec:individuals}

In the previous section, evidence for systematic differences between groups was considered, while allowing for the presence of individual variability.  This section considers situations where interest lies in the evaluation of individuals.  Traits which can be expressed in single values are considered, as well as more general characterisation of the particularities of individual shapes.

\subsection{Asymmetry}

For shapes whose ideal form is symmetric, deviations which disturb this symmetry are important features.  The left/right symmetry of human faces is a major example, where any strong departure from symmetry creates a striking visual impression.  However, real faces are all asymmetric to some degree so, as part of the process of evaluating an individual shape, it is important to characterise the asymmetries found in an appropriate reference population.

When a shape is represented by a set of point locations, some of which are paired as left/right counterparts, quantification of asymmetry is generally based on the degree of post-registration mismatch between the shape and its reflection with the left/right labels swapped.  Theoretical development of this idea was undertaken by \citet{mardia-2000-biometrika} and \citet{kent-2001-biometrika} in the context of landmarks, and many authors have exploited this thinking in biological contexts.  \citet{bock-2006-applstat} proposed a decomposition of global asymmetry which allowed local sources to be identified and separated into contributions from individual features and their configurations.

The first step in computing a functional measure of asymmetry for a surface $X(s)$ is to apply (functional) Procrustes matching of the mirror image onto the original surface, to create the new surface $\tilde{X}(s)$.  The mirror image is created in practice by reflecting and relabelling the configuration of points which express the shape model.  The integrated comparison and its discrete approximation are then easily constructed as
$$
      \frac{1}{A(\mathcal{S})} \int_\mathcal{S} ||X(s) - \tilde{X}(s)||^2 ds
      \approx
      \frac{1}{\left(\sum_{j=1}^J a_j\right)} \sum_{j=1}^J ||X(s_j) - \tilde{X}(s_j)||^2 a_j ,
$$
where, to be even-handed, $\mathcal{S}$ is the surface formed from the average of $X(s)$ and $\tilde{X}(s)$, $A(\mathcal{S})$ denotes its surface area and, as usual, the $a_j$ give the areas surrounding the individual triangles.  The final asymmetry score is achieved by applying a square root transformation, so that the scale of the end result matches the scale of the original co-ordinate measurements.

\begin{figure}
\begin{minipage}{0.46\textwidth}
   \includegraphics[width = 0.5\textwidth]{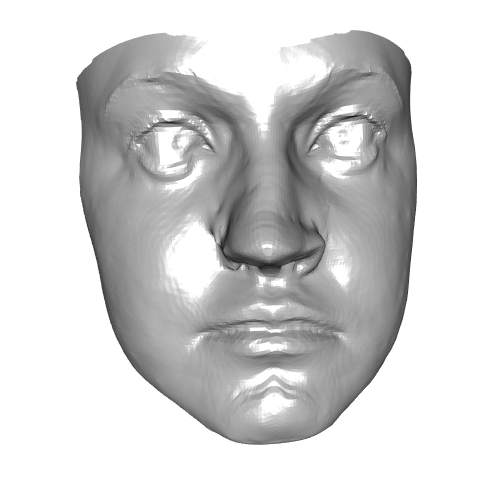}
   \hspace*{-5mm}
   \includegraphics[width = 0.5\textwidth]{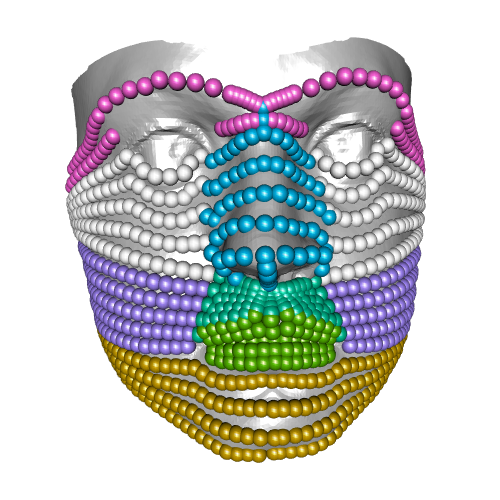} \\
   \includegraphics[width = 0.5\textwidth]{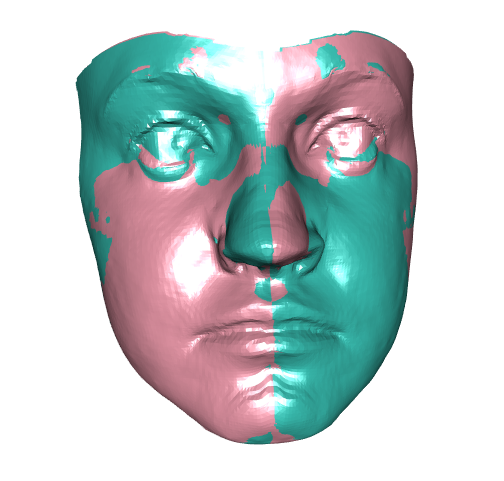}
   \hspace*{-5mm}
   \includegraphics[width = 0.5\textwidth]{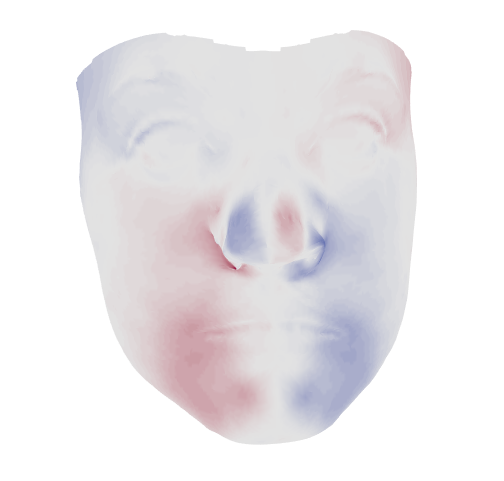} \\
   \hspace*{8mm}
   \includegraphics[width = 0.1\textwidth, angle = -90]{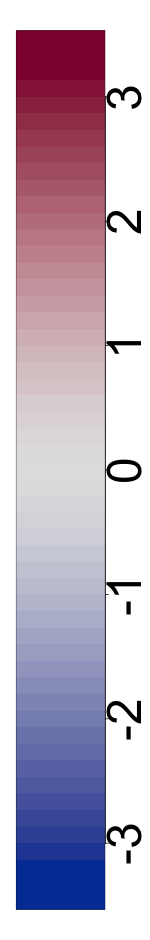}
\end{minipage}
\begin{minipage}{0.52\textwidth}
   \includegraphics[width = \textwidth]{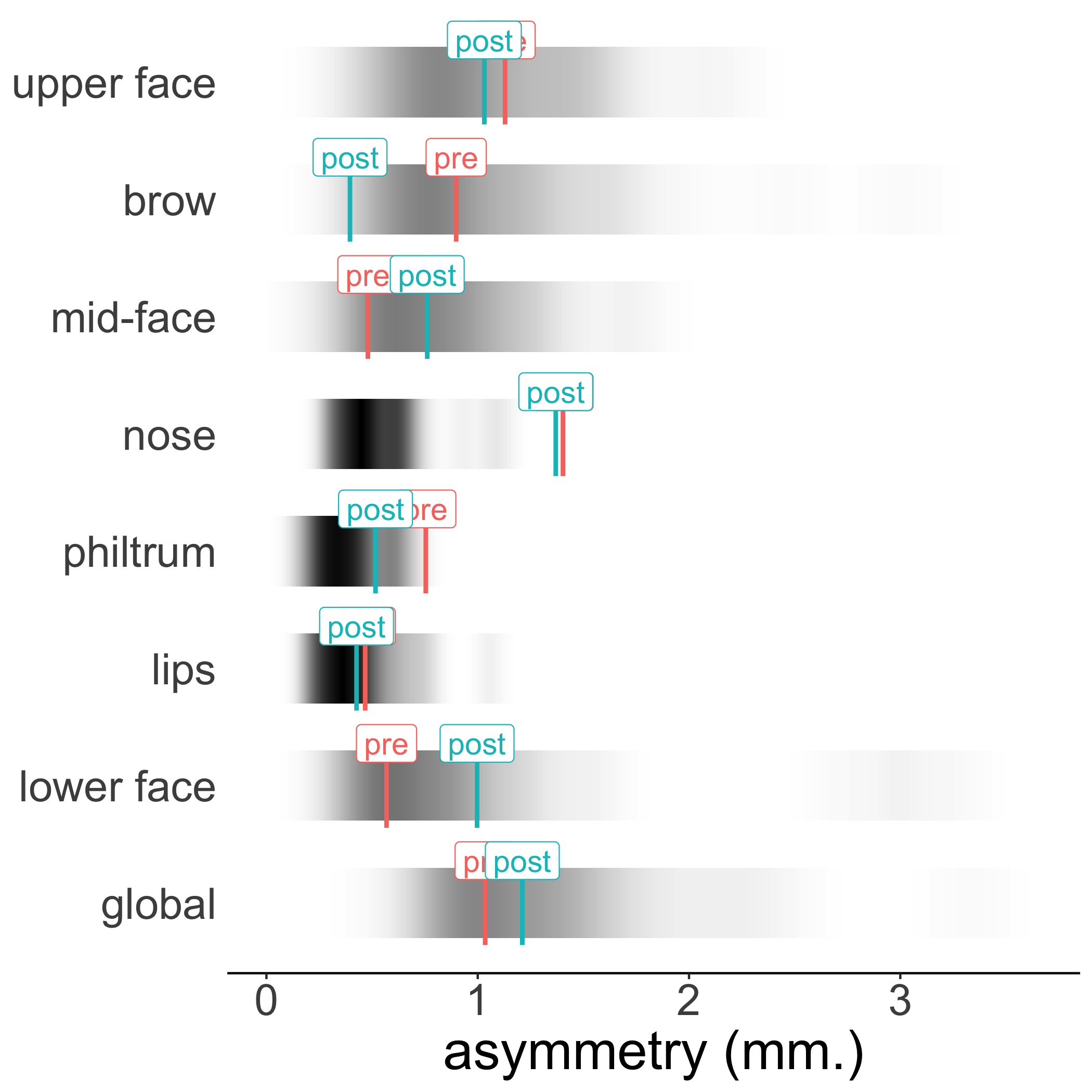}
\end{minipage}
\caption{The four facial images show, in clockwise order from top right, a post-surgical patient, the sub-regions used to compute asymmetry scores, the colour-coded distance between the shape and its matched reflection, and the superimposition of the shape and its matched reflection.  The density strips show the asymmetry scores from control faces, with the pre- and post-surgical scores for the individual superimposed.}
\label{fig:asymmetry}
\end{figure}

An example of asymmetry scores in action is provided by orthognathic surgery, where the maxilla or mandible of a patient is repositioned to improve the alignment of teeth and to address issues of facial appearance.  \citet{vittert-2018-ijoms} included asymmetry in their assessment of post-surgical outcome, identifying evidence of a reduction in mean asymmetry in the upper lip region.  However, asymmetry scores also give the opportunity to assess patients individually.  Figure~\ref{fig:asymmetry} shows the facial image of one post-surgical patient together with comparison of the reflected and matched image in both superimposition and colour-coded forms.  Interpretation of this information is informed by quantifying the asymmetry scores exhibited in an adult control population, with the distributions represented in the right hand side of Figure~\ref{fig:asymmetry} by density strips \citep{jackson-2008-amerstat}.  The global asymmetry scores, for both pre-surgical and post-surgical facial shapes of this patient, have been superimposed on the bottom density strip.  These scores are entirely typical of controls and in particular they provide reassurance that surgery has not introduced any marked asymmetry overall.  The scores have also been computed for a variety of sub-regions, indicated by the top right hand image of the four facial images.    The scores and density strips indicate strong nasal asymmetry, but this is apparent both before and after surgery and so it cannot be attributed to surgical intervention.

\subsection{Closest Controls}

For more general assessment of the shapes of individual cases, an approach analogous to the concept of a `normal range' for univariate data is required.  In the surgical context, characterising any differences between a post-surgical patient and a control population could provide helpful guidance on the nature of any further surgery which may be required. \citet{bowman-2006-jcgs} outlined an approach based on the concept of a `closest control'.  This identifies the shape which is as close as possible to the individual of interest but which lies on the surface of a $95$\% prediction ellipsoid and so lies within the `normal range' associated with controls.  Any remaining shape differences then characterise the features of the individual shape which are different from controls.

\citet{bowman-2006-jcgs} derived the algebra of this in a simple case involving curve data, using a principal component regularisation to reduce dimensionality across both cases and controls.  The concept is applied here to surface data but the ideas are developed further in two important ways.  Firstly, principal components are constructed from the control data only.  This gives a clear interpretation of the components which is unaffected by the particular cases available.  Secondly, variation unexplained by these components is also considered, in order to give a complete description of the observed data.

If a new shape $Z$, such as a post-surgical patient, is registered onto the control mean then it can be projected into the space of the first $p$ principal components, denoted by $\mathcal{C}_p$,  by computing the score vector $v(Z) = \mbox{vec}(Z - \bar{X})^T E_p$, where $E_p$ is the matrix whose columns contain the first $p$ principal component vectors derived from the control data.  Within this space, the Mahalanobis distance of the new shape from the mean control shape is
$$
	d(Z) = v(Z)^T \hat{\Sigma}^{-1} v(Z)
$$
where $\hat{\Sigma}$ is a diagonal matrix containing the variances of the principal components.  The Mahalanobis distance has a $\chi^2_p$ distribution approximately.  If $d(Z)$ is less than the $95$th percentile of this distribution, denoted by $\chi^2_p (0.95)$, then the new shape falls within the `normal range' of controls in this space.  If $d(Z) > \chi^2_p (0.95)$ then the \textit{closest control} in this $p$-dimensional space can be found by shrinking $v$ towards $0$ until its Mahalanobis distance matches $\chi^2_p (0.95)$.  The shrinking factor $\alpha_1$ is easily found as $\alpha_1 = \sqrt{\frac{\chi^2_p (0.95)}{d(Z)}}$, by solving the equation $\alpha_1 v(Z)^T \hat{\Sigma}^{-1} \alpha_1 v(Z) = \chi^2_f (0.95)$.  The scores of this new location $\alpha_1 v(Z)$ are then converted into tangent co-ordinates as $\alpha_1 v(Z) E_p^T$, and expressed as a shape by reconfiguring the tangent co-ordinates into a three column matrix in the usual manner as
$$
      cc_p(Z) = \bar{X} + \mbox{vec}^{-1}\{\alpha_1 v(Z) E_p^T\} .
$$

This finds the closest control in $\mathcal{C}_p$.  However, the case of interest may well have shape features which cannot be captured in this space so characterisation in the complementary space, denoted by $\mathcal{R}_p$, is also required.  The projection of $Z$ onto $\mathcal{C}_p$ is
$$
      \tilde{Z} = \mbox{vec}^{-1} \left\{ v(Z)  E_p^T \right \} ,
$$
so the relevant information is found in the residual shape $R(Z) = Z - \tilde{Z}$.  The length of the residual at each model location can be quantified in the vector $L(Z) = \sqrt{R(Z)^2 1_3}$, where here the square-root and the exponent $2$ are applied element-wise.  A measure of variation in the lengths of the residuals at each model location for controls is then available in the vector $\nu$ whose $j$th element is the standard deviation of $\{L(X_i)_j; i = 1, \ldots, n\}$, where $L(X_i)_j$ denotes the $j$th element of $L(X_i)$.  A simple measure of variation is then
$$
      r(Z) = \frac{1}{J} \sum_{j = 1}^J L(Z)_j / \nu_j .
$$
This averages the lengths of the residuals across the model locations, standardised at each model location by the variation in control residual length.  The value of $r(Z)$ may be regarded as atypical if it lies beyond $q_{95}$, the $95$ quantile of $\{v(X_i); i = 1, \ldots, n\}$.  A closest control in the residual space, $\mathcal{R}_p$, can then be constructed by shrinking the residual shape to $\alpha_2 Z$, where $\alpha_2 = q_{95} / r(Z)$.  An overall closest control for $z$ can now be constructed as
$$
      cc(Z) = cc_p(Z) + \alpha_2 R(Z)
$$
which combining the closest controls in the sub-spaces $\mathcal{R}_p$ and $\mathcal{C}_p$.

\begin{figure}
\centerline{
   \includegraphics[width = 0.5\textwidth]{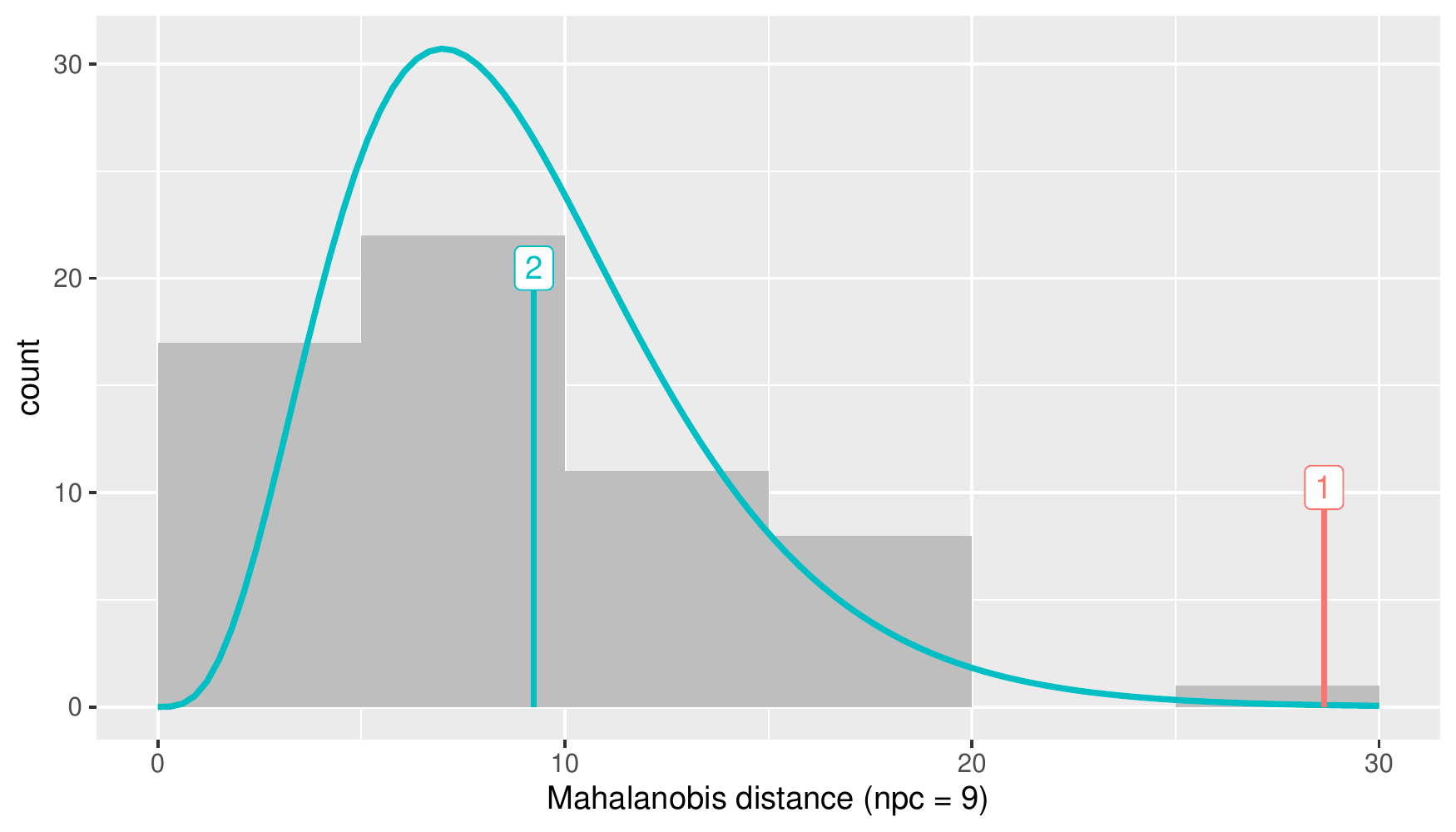}
   \includegraphics[width = 0.5\textwidth]{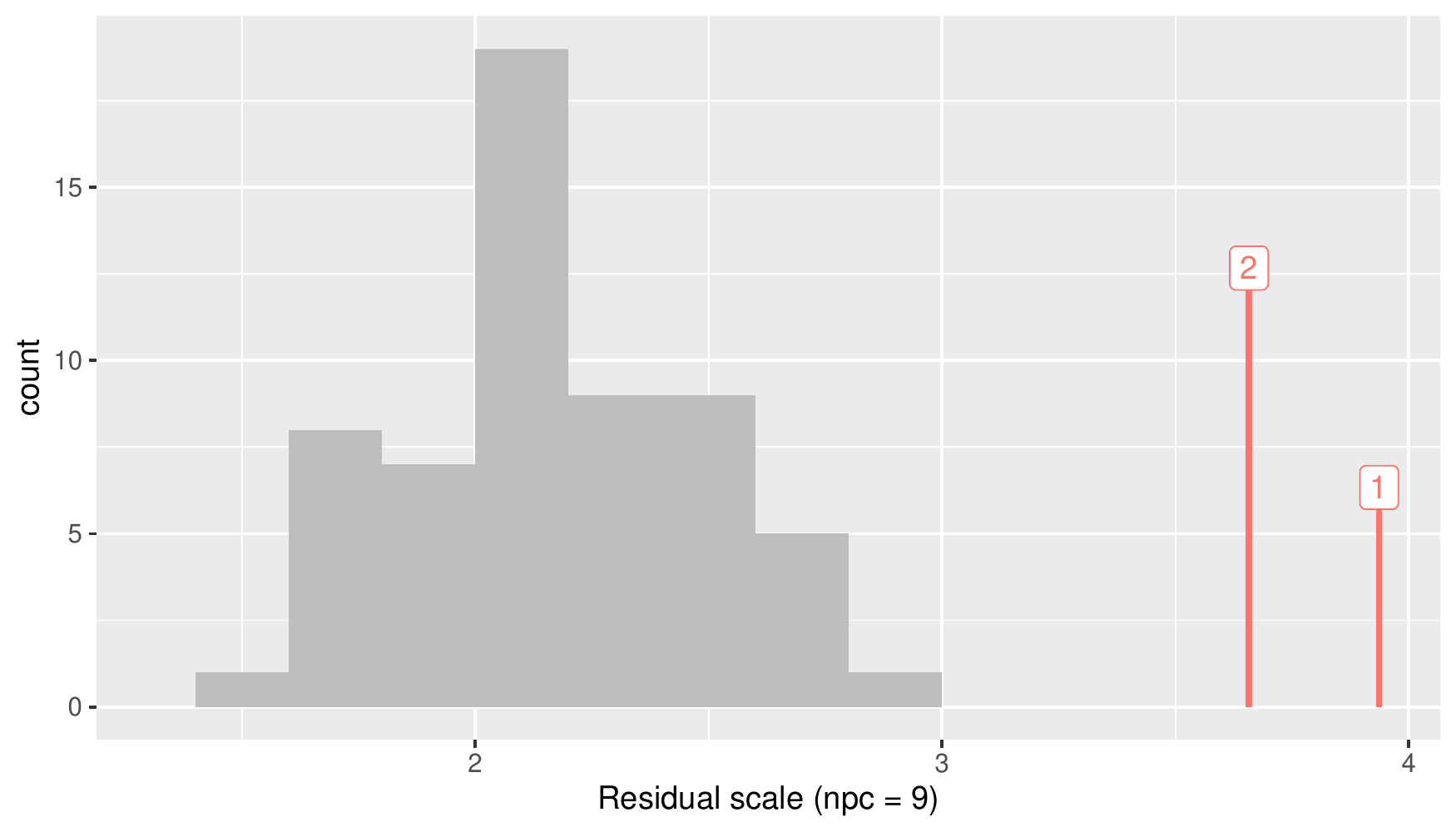}
   }
\centerline{
   \includegraphics[width = 0.25\textwidth]{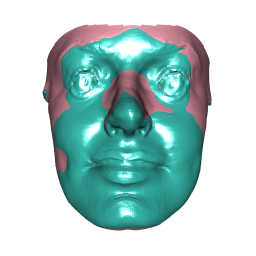}
   \includegraphics[width = 0.25\textwidth]{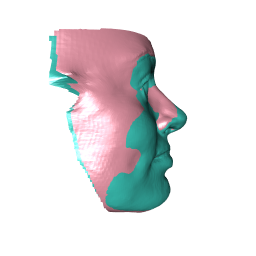}
   \includegraphics[width = 0.25\textwidth]{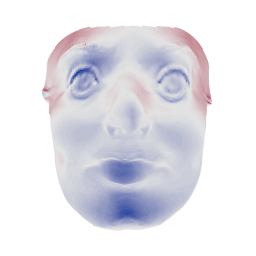}
   \includegraphics[width = 0.04\textwidth]{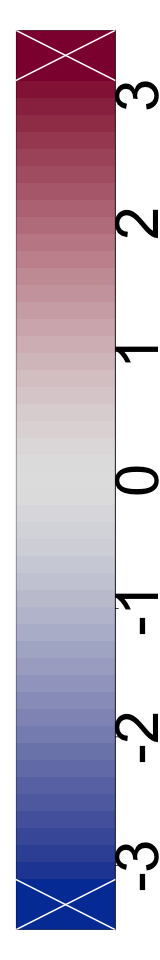}
}
\caption{The histograms show the distances of the control shapes from the mean in the space of the first $9$ principal components (left) and in the residual space (right).  The distances of two post-surgical cases from the control mean are superimposed.  The lower images compare the facial shape of case number 1 (green) with its closest control. both through superimposition and as normal distances from case to closest control.}
\label{fig:cc-orthognathic}
\end{figure}

Figure~\ref{fig:cc-orthognathic} shows the results of applying the concept of closest control to two post-surgical orthognathic cases.  The left-hand histogram shows the Mahalanobis distances of controls in $\mathcal{C}_p$, the space of the first $p$ principal components for controls.  The use of $p = 9$ was determined by the smallest number of components which explained at least $80$\% of the variation in the controls.  Case $1$ clearly lies in the tails of the control distribution while case $2$ in unexceptional.  The right-hand histogram shows that both cases exhibit unusual behaviour in the residual space, $\mathcal{R}_p$.  However, shape differences in this residual space may be small.  The facial images in the lower part of Figure~\ref{fig:cc-orthognathic} compare case $1$ (green) with its the closest control (pink) by superimposition and by normal distances.  This characterises the unusual features of the case as a slightly more prominent lower face than in controls, particularly in the mandible (lower jaw).  This is potentially valuable feedback on surgery which involves repositioning of the underlying bones.  A display of the closest control information for case $2$ is deferred to the next sub-section.

\subsection{An integrated patient assessment}
\label{subsec:refcard}

The methods described in this section provide valuable tools for the characterisation of individual shapes of interest.  The combination of these tools forms the basis of an integrated patient assessment.  This is illustrated in Figure~\ref{fig:scorecard}, using case $2$ from Figure~\ref{fig:cc-orthognathic}.  This brings together the observed pre-surgical and post-surgical shapes, comparisons of this case with control shape both for closest control analysis and for asymmetry, and illustrates differences in shape through superimposition and normal distances.  An interactive display would allow those reviewing the case to inspect the shapes in 3D and to query further information.  However, this static display gives a helpful summary of the effects of surgery on this particular patient.

\definecolor{grey}{gray}{0.85}
\newcommand{\greybox}[1]{
   \fcolorbox{black}{grey}{
      \parbox{\textwidth}{
         \textcolor{black}{\centerline{\textsf{#1}}}
      }
   }
}

\begin{figure}
\begin{minipage}{0.85\textwidth}
\greybox{Pre-surgical shape \hspace{3cm} Post-surgical shape} \\
   \includegraphics[width = 0.24\textwidth]{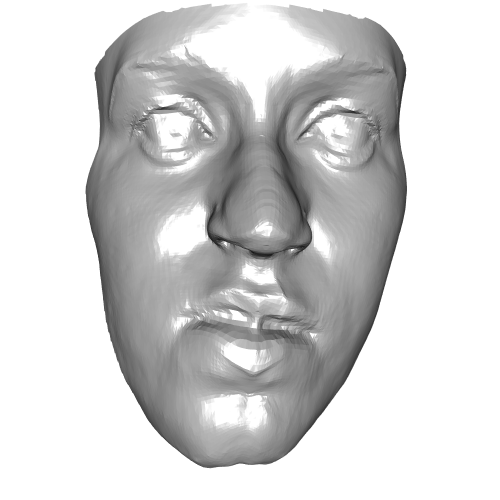}
   \includegraphics[width = 0.24\textwidth]{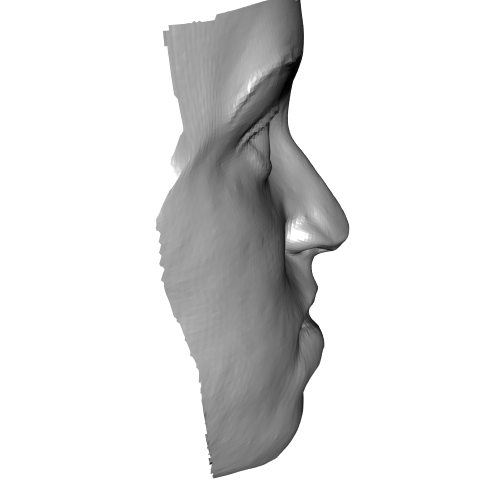}
   \includegraphics[width = 0.24\textwidth]{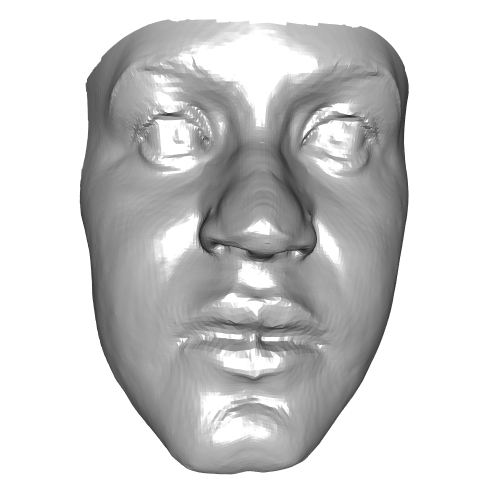}
   \includegraphics[width = 0.24\textwidth]{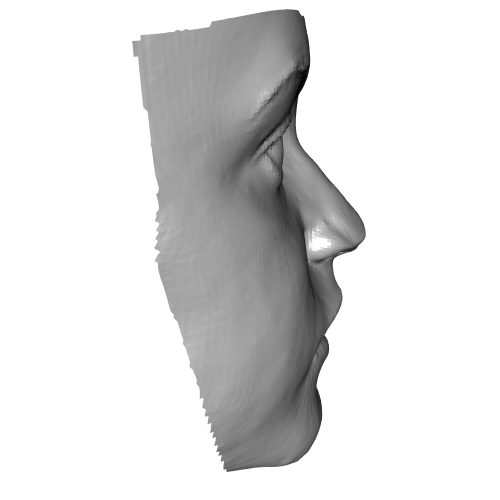}
   
\greybox{Closest control} \\
   \begin{minipage}{0.24\textwidth}
      {\small \hspace{4mm} \textsf{Pre-surgery}} \\
      \includegraphics[width = \textwidth]{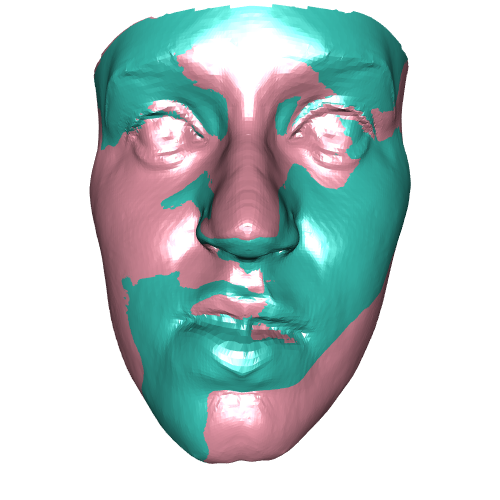}
      \includegraphics[width = \textwidth]{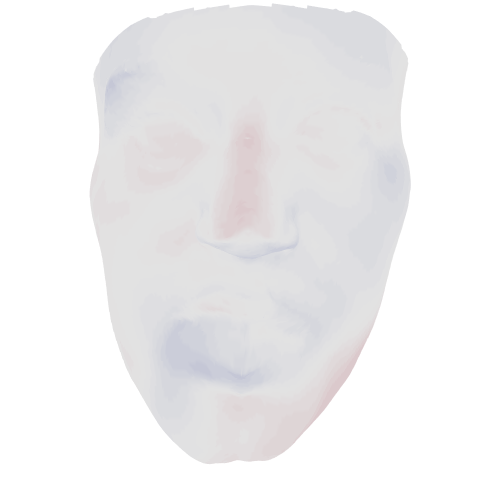}
   \end{minipage}
   \begin{minipage}{0.48\textwidth}
      \includegraphics[width = \textwidth]{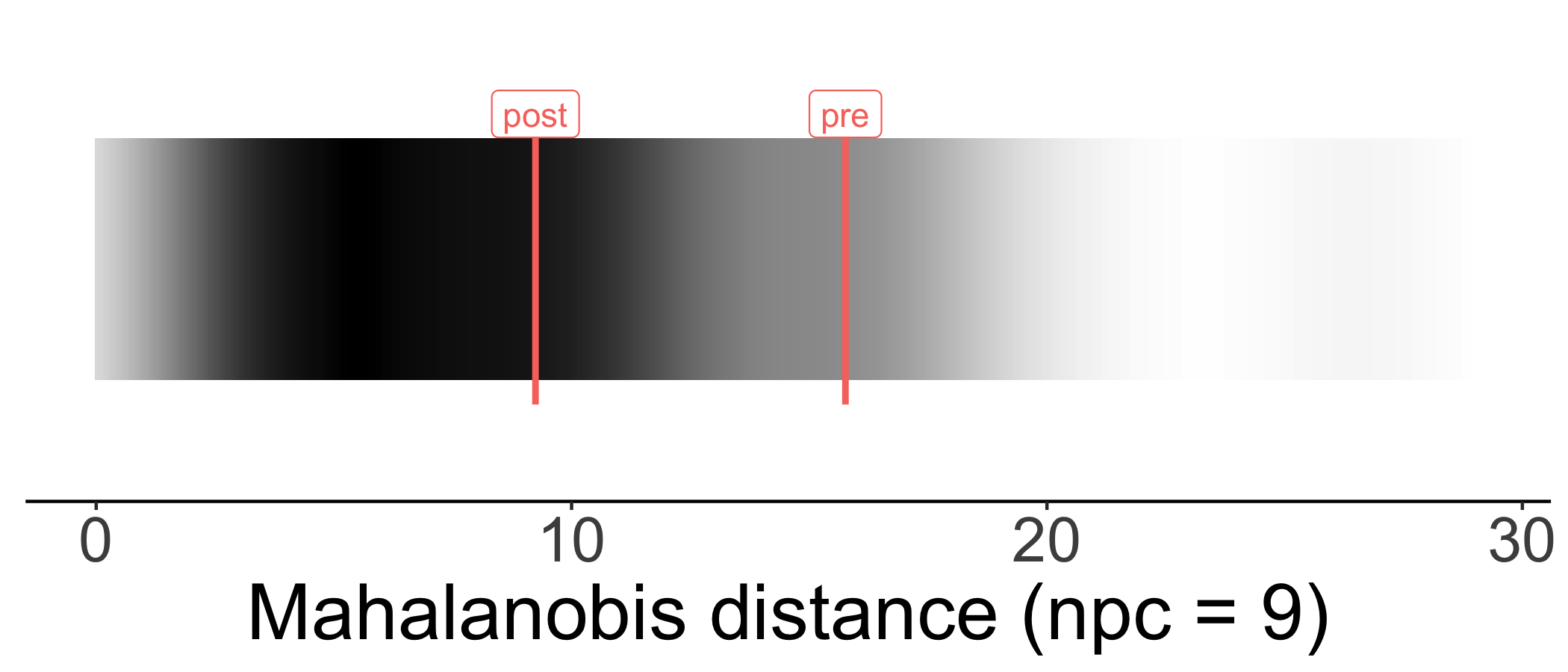}
      \includegraphics[width = \textwidth]{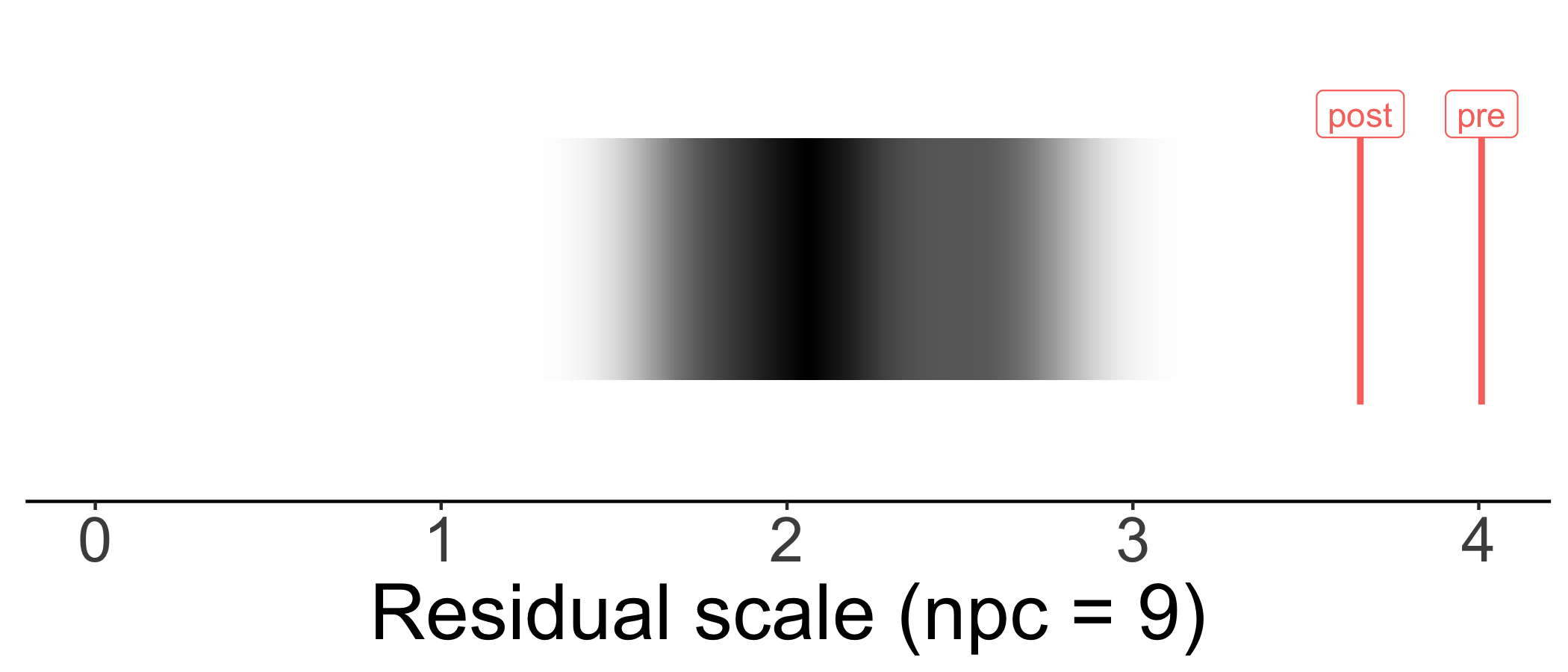}
   \end{minipage}
   \begin{minipage}{0.24\textwidth}
      {\small \hspace{4mm} \textsf{Post-surgery}} \\
      \includegraphics[width = \textwidth]{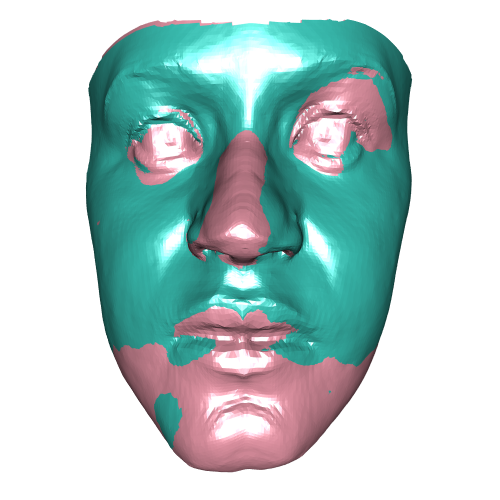}
      \includegraphics[width = \textwidth]{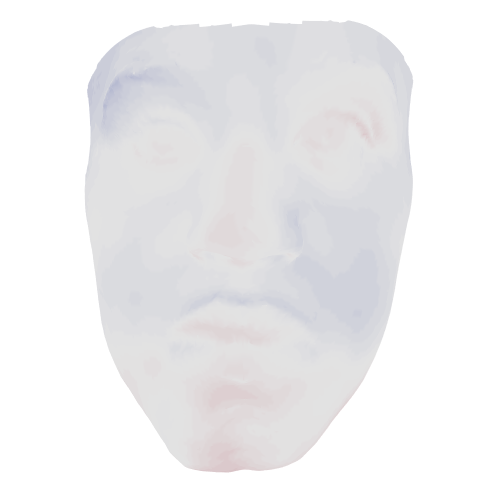}
   \end{minipage}

\greybox{Asymmetry} \\
   \begin{minipage}{0.24\textwidth}
      {\small \hspace{4mm} \textsf{Pre-surgery}} \\
      \includegraphics[width = \textwidth]{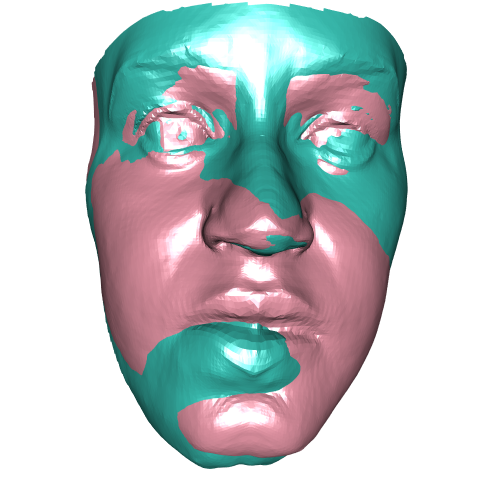}
      \includegraphics[width = \textwidth]{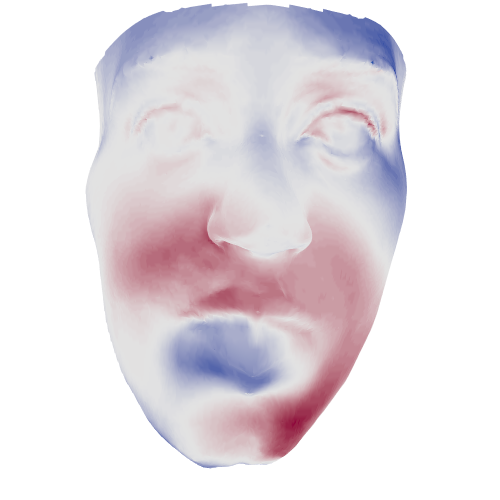}
   \end{minipage}
   \begin{minipage}{0.48\textwidth}
      \includegraphics[width = \textwidth]{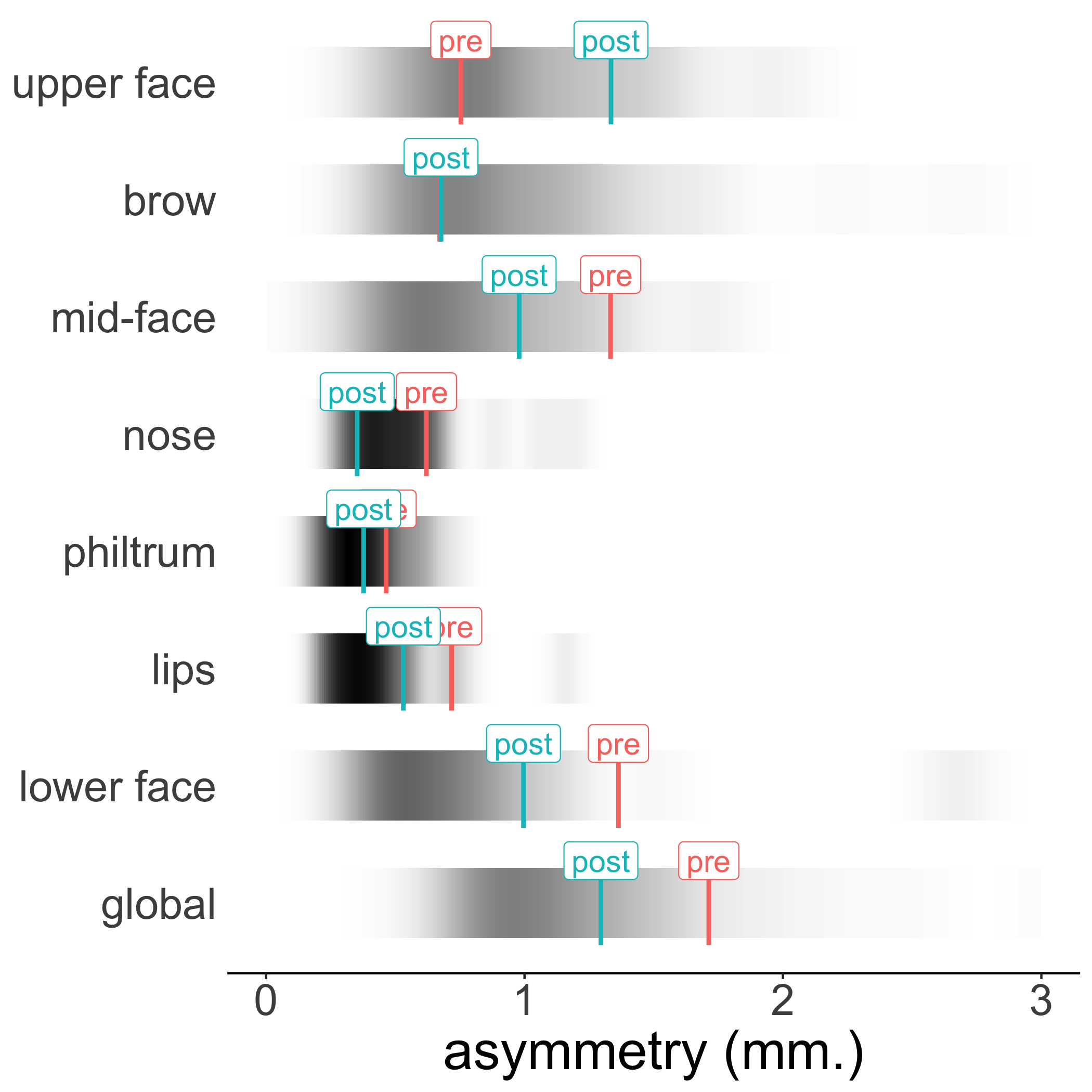}
   \end{minipage}
   \begin{minipage}{0.24\textwidth}
      {\small \hspace{4mm} \textsf{Post-surgery}} \\
      \includegraphics[width = \textwidth]{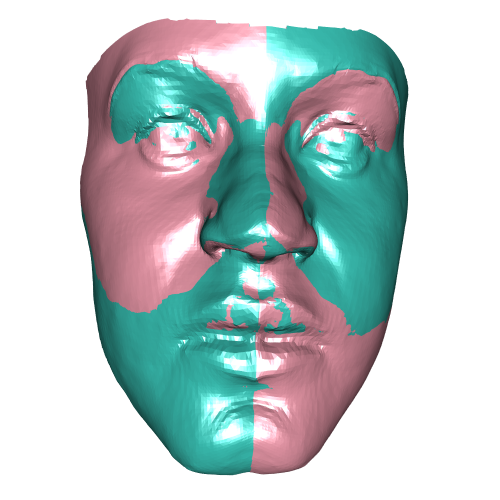}
      \includegraphics[width = \textwidth]{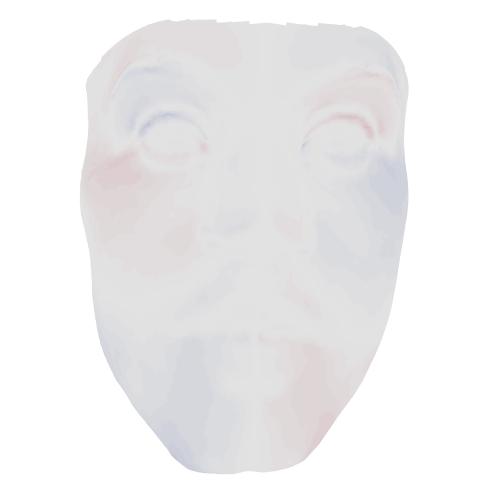}
   \end{minipage}
\end{minipage}
~~
\begin{minipage}{0.09\textwidth}
      ~ \\ ~ \\ ~ \\ ~ \hspace*{0.5mm} \textsf{\small mm} \\
      \includegraphics[width = \textwidth]{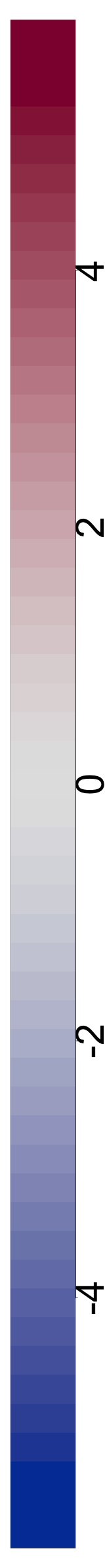}
\end{minipage}
\caption{An integrated assessment of patient $2$ from Figure~\ref{fig:cc-orthognathic}.}
\label{fig:scorecard}
\end{figure}


\section{Discussion}
\label{sec:discussion}

This paper has proposed methods of analysis for high resolution surface data and  corresponding models which give consistent descriptions of each observed shape.  A strong emphasis has been on the adoption of functional forms of analysis and the difficulty of identifying a common sample space has been overcome by using the mean surface as an indexing shape.  This enabled functional forms of registration, principal components analysis and group comparisons to be developed.  In applying these methods, strong emphasis was also placed on the use of principal components to identify sub-spaces of interest rather than inspection of individual components.  Graphical displays of shape change which jointly describe these sub-spaces were also adopted.  Particular attention was given to the comparison of individual shapes with a relevant control group.  In additional to univariate measures such as asymmetry scores, the concept of a `closest control' was developed in detail, to give a powerful means of identifying any unusual characteristics of an individual shape of interest.  

These methods provide powerful tools for the assessment of surface shapes and the practical implications of their use, particularly in surgical contexts, will be the subject of subsequent research.


\section*{Acknowledgement}

The early stages of the research carried out by Adrian Bowman and Stanislav Katina was supported by a Wellcome Trust grant (WT086901MA) to the Face3D research consortium (\texttt{www.Face3D.ac.uk}), under whose auspices the facial data were collected.


\section{Supplementary materials}

Animated version of some of the Figures in the paper are available through the links below.

\begin{description}

\item[Figure~\ref{fig:sex-means-colour}:] ~ \\
\url{http://www.stats.gla.ac.uk/~adrian/JCGS/FtoM-frontal.mp4} \\
\url{http://www.stats.gla.ac.uk/~adrian/JCGS/FtoM-lateral.mp4}

\item[Figure~\ref{fig:grand-tour}:] ~ \\
\url{http://www.stats.gla.ac.uk/~adrian/JCGS/F-1-frontal.mp4} \\
\url{http://www.stats.gla.ac.uk/~adrian/JCGS/F-1-lateral.mp4} \\
\url{http://www.stats.gla.ac.uk/~adrian/JCGS/F-2-frontal.mp4} \\
\url{http://www.stats.gla.ac.uk/~adrian/JCGS/F-2-lateral.mp4} \\
\url{http://www.stats.gla.ac.uk/~adrian/JCGS/F-3-frontal.mp4} \\
\url{http://www.stats.gla.ac.uk/~adrian/JCGS/F-3-lateral.mp4} \\
\url{http://www.stats.gla.ac.uk/~adrian/JCGS/F-4-frontal.mp4} \\
\url{http://www.stats.gla.ac.uk/~adrian/JCGS/F-4-lateral.mp4} \\
\url{http://www.stats.gla.ac.uk/~adrian/JCGS/tour.mp4}

\end{description}

\newpage
\begin{appendices}

\section{3D warping}

The technical details of warping are described here because the method is not widely used in 3D.  We seek a function which maps $X$ onto $Y$ exactly.  If an interpolant of a single co-ordinate of $Y$ as a function of the three co-ordinates of $X$ is considered, then the elegant functional analysis described by \citet{duchon-1977-book} provides an immediate solution. The aim is to find the interpolating function $f$ which has minimal bending energy, defined as
$$
      \int_{\mathcal{R}^3} \left\{ \sum_{p=1}^3 \sum_{q=1}^3 \left( \frac{\delta^2f}{\delta x_p \delta x_q}\right)^2 \right\} dx_1 dx_2 dx_3
.$$
The solution can be expressed in terms of radial basis functions which parameterise the relationship between points $x$ and $y$ in $\mathbb{R}^3$ as
$$
      y_d(x) = \sum_{j=1}^J \phi\left(||x - x_j||\right) \beta_{jd} ,
$$
where $\beta_{jd}$ are parameters and $d$ denotes the three dimensions of $\mathbb{R}^3$.  Fitting this functional form to the mapping from the observed locations in $X$ to those in $Y$ requires $Y = S \beta_1$, where $S$ is a $J \times J$ matrix, with $S_{ij} = \phi\left(||x_i - x_j||\right)$, and $\beta_1$ is a $J \times 3$ matrix whose $(j,d)$th element is $\beta_{jd}$.

It is helpful to separate the mapping into affine and non-affine components, with the former capturing the linear part of the transformation, including possibly different scalings in different co-ordinate directions (shear), and the latter describing non-linear bending.  If $Q$ denotes the matrix $(1_J \ X)$, where $1_J$ is a column vector of $1$'s, then the transformation can be written in the multivariate form
$$
      Y = S \beta_1 + Q \beta_2 ,
$$
where $\beta_2$ is a $4 \times 3$ matrix filled with parameters.  This system is now over-parametrised, with $(J + 4) \times 3$ parameters but only $J \times 3$ defining equations.  This can easily be resolved by adopting suitable constraints, for example through the extended system
\begin{equation}
      \left( \begin{array}{c} Y \\ 0 \end{array} \right) = 
            \left( \begin{array}{cc} S & Q \\ Q^T & 0 \end{array} \right)
            \left( \begin{array}{c} \beta_1 \\ \beta_2 \end{array} \right) ,
\label{eq:warping}
\end{equation}
where the $0$ entries indicate matrices filled with $0$'s of the dimensionality required by the context.  These constraints require the sum of the entries of each column of $\beta_1$ to be $0$ and the sum weighted by the co-ordinates of each dimension of $X$ also to be $0$.  By applying constraints to the affine component, the interpretation of  the non-affine component is left undisturbed.  

The system of equations (\ref{eq:warping}) can be written in the condensed form $Y_e = X_e \beta$, with obvious definitions of $X_e$ and $Y_e$.  If the matrix $S$ is invertible then so is $X_e$ and, after some standard matrix manipulations, the solutions emerge as
\begin{eqnarray*}
      B_e      & = & \left( S^{-1} - S^{-1} Q \left(Q^T S^{-1} Q \right)^{-1} Q^T S^{-1} \right) , \\
      \beta_1 & = & B_e Y , \\
      \beta_2 & = & \left(Q^T S^{-1} Q \right)^{-1} Q^T S^{-1} Y .
\end{eqnarray*}
When the \textit{bending energy} matrix $B_e$ is post-multiplied by $X$, this generates the coefficients of the non-affine part of the transformation.  The bending energy itself can be expressed as $\mbox{tr}\left\{Y^T B_e Y\right\}$. Finally, the optimal radial basis function is shown simply to be $\phi(z) = -\frac{1}{8\pi} z$.

\end{appendices}

\bibliographystyle{Chicago}
\bibliography{shape-visualisation-paper.bib}

\end{document}